\documentclass[]{macro/neu_msthesis}
\usepackage{tabularx}
\usepackage{caption}
\usepackage{subcaption}
\usepackage{comment}
\usepackage{siunitx}
\usepackage{float}
\usepackage{graphics}
\usepackage{lipsum}
\usepackage{color}

\usepackage[backend=biber,style=numeric,sorting=none]{biblatex}
\DeclareSourcemap{
  \maps{
    \map{
      \pertype{article}
      \step[fieldset=language, null]
      \step[fieldset=url, null]
      \step[fieldset=doi, null]
      \step[fieldset=issn, null]
      \step[fieldset=isbn, null]
      \step[fieldset=note, null]
      \step[fieldset=editor, null]
      \step[fieldset=urldate, null]
      \step[fieldset=file, null]
      \step[fieldset=eprint, null]
    }
  }
}
\DeclareSourcemap{
  \maps{
    \map{
      \pertype{book}
      \step[fieldset=language, null]
      \step[fieldset=url, null]
      \step[fieldset=doi, null]
      \step[fieldset=issn, null]
      \step[fieldset=isbn, null]
      \step[fieldset=note, null]
      \step[fieldset=editor, null]
      \step[fieldset=urldate, null]
      \step[fieldset=file, null]
      \step[fieldset=eprint, null]
    }
  }
}
\DeclareSourcemap{
  \maps{
    \map{
      \pertype{inproceedings}
      \step[fieldset=language, null]
      \step[fieldset=url, null]
      \step[fieldset=doi, null]
      \step[fieldset=issn, null]
      \step[fieldset=isbn, null]
      \step[fieldset=note, null]
      \step[fieldset=editor, null]
      \step[fieldset=urldate, null]
      \step[fieldset=file, null]
    }
  }
}
\DeclareSourcemap{
  \maps{
    \map{
      \pertype{incollection}
      \step[fieldset=language, null]
      \step[fieldset=url, null]
      \step[fieldset=doi, null]
      \step[fieldset=issn, null]
      \step[fieldset=isbn, null]
      \step[fieldset=note, null]
      \step[fieldset=editor, null]
      \step[fieldset=urldate, null]
      \step[fieldset=file, null]
    }
  }
}
\DeclareSourcemap{
  \maps{
    \map{
      \pertype{unpublished}
      \step[fieldset=language, null]
      \step[fieldset=url, null]
      \step[fieldset=doi, null]
      \step[fieldset=issn, null]
      \step[fieldset=isbn, null]
      \step[fieldset=note, null]
      \step[fieldset=editor, null]
      \step[fieldset=urldate, null]
      \step[fieldset=file, null]
      \step[fieldset=eprint, null]
      \step[fieldset=primaryclass, null]
    }
  }
}
\DeclareSourcemap{
  \maps{
    \map{
      \pertype{misc}
      \step[fieldset=language, null]
      \step[fieldset=url, null]
      \step[fieldset=doi, null]
      \step[fieldset=issn, null]
      \step[fieldset=isbn, null]
      \step[fieldset=note, null]
      \step[fieldset=editor, null]
      \step[fieldset=urldate, null]
      \step[fieldset=file, null]
    }
  }
}
\usepackage[usenames,dvipsnames]{xcolor}
\usepackage{tcolorbox}
\usepackage{array}
\usepackage{colortbl}
\usepackage{booktabs}
\usepackage{enumerate}

\title{Legged Walking on Inclined Surfaces}

\author{Chenghao Wang}

\nuid{002199930}

\dept{Electrical and Computer Engineering}

\degreename{ECE - Concentration Hardware and Software for Machine Intelligence }

\field{ECE - Concentration Hardware and Software for Machine Intelligence }

\submitdate{April 2022}

\numberofmembers{3}
\principaladviser{Dr. Alireza Ramezani}
\chairman{Dr. Miriam Leeser}
\dean{Dr. Waleed Meleis}
\usepackage{amsfonts,amssymb,amsmath}			% special math characters
\usepackage{times}		% use Postscript fonts
\usepackage{bm}

\newcommand{\ifno}[1]{}

\usepackage{multirow}
\makeindex
\usepackage{makeidx}
\usepackage{acronym}

\usepackage{url}
\ifnum\pdfoutput>0
\usepackage{graphicx}
\else
\usepackage{graphicx}
\fi

\ifnum\pdfoutput>0
\usepackage[pdftex,%                    % hyper-references for ps2pdf
bookmarks=true,%                        % generate bookmarks ...
bookmarksnumbered=true,%                % ... with numbers
hypertexnames=false,%                   % needed for correct links to figures
breaklinks=true           %Allows link text to break across lines;
]{hyperref}
\else
\usepackage[hypertex,%                  % hyper-references for ps2pdf
bookmarks=true,%                        % generate bookmarks ...
bookmarksnumbered=true,%                % ... with numbers
hypertexnames=false,%                   % needed for correct links to figures
breaklinks=true           %Allows link text to break across lines;
]{hyperref}
\fi

\hypersetup{				% setup PDF information fields
pdfauthor = {\authorRef},
pdftitle = {\titleRef},
pdfsubject = {\expandafter{\degreeRef} thesis submitted to Northeastern University},
pdfkeywords = {add keywords here}
pdfcreator = {LaTeX with hyperref package},
pdfproducer = {dvips + ps2pdf}}

\usepackage{microtype}

\begin{document}

% add a pdf bookmark to the cover page
\pdfbookmark[1]{Cover}{cover}

% --- title page ---
\titlepage

% --- front matter ---
\begin{frontmatter}
% %%%%%% NO SIGNATURE PAGE WILL BE GENERATED, PLEASE DOWNLOAD FROM COE WEB SITE (READ file Readme.pdf in this folder!)
%\signaturepage
% dedication

% dedication.tex:

\begin{dedication}
To my family.
\end{dedication}

% table of content (add bookmark for convenience)

\pdfbookmark[1]{Table of Contents}{contents}
\tableofcontents
\listoffigures
\newpage\ssp
\listoftables

% include a list of Acronyms (comment out if no acronyms are specified)
% acronyms.tex
\chapter*{List of Acronyms}
\addcontentsline{toc}{chapter}{List of Acronyms}

% below is the list of acronym definitions, place them in alphabetical order
% since they will not be sorted again. 
\begin{acronym}
\item DoF - Degrees of Freedom.

\item HROM - Husky Reduced Order Model.

\item LTI - Linear Time Invariant.

\item LQR - Linear Quadratic Regulator.

\item MPC - Model Predictive Control.

\item GRF - Ground Reaction Forces.

\item HZD - Hybrid Zero Dynamics.

\item ZMP - Zero Moment Point.

\item CoM - Center of Mass.

\item CoP - Center of Pressure.

\item HF - Hip Frontal.

\item HS - Hip Sagittal.

\end{acronym}

%% include any of the front matter files that contain text
%% attention the input does cause a page break, the include on
%% the other hand does not
% acknowledgements.tex:

\begin{acknowledgements}
In this thesis, I need to express my gratitude to my advisor, Prof. Alireza Ramezani, for his persistent guidance throughout my entire master's journey. I would like to extend my appreciation to my thesis committee member Prof. Leeser and Prof. Shafai for their detailed suggestions on the thesis and attending my defense. And I need to thank Mingxi Jia for his support over the years. His enthusiasm for research and optimism about life have continued motivating me. I would also like to thank all of my lab mates, including Kaushik, Shreyansh, Adarsh, Shoghair, Bibek, Xuejian, Yizhe, and Xintao, for their help with hardware, experiments, control design, simulation, and consistent encouragement.
\end{acknowledgements}

% abstract.tex:

\begin{abstract}

% Legged locomotion has been a field of high interest and significant advancements with various quadrupedal platforms developed deploying successful controllers for walking, trotting, and hopping robustly. One area still not highly explored is integrating the abilities of legged and aerial robots into a single platform and evaluating the legged locomotion capabilities of such systems. This thesis explores the control of a custom legged and aerial locomotion platform, Husky Carbon. Husky carbon is an 18 Degree of Freedom(DoF) quadruped capable of morphing into quad-copter configuration for aerial mobility. The design of the system for both modes of operation brings novel considerations for the implementation of gaits and achieving successful legged locomotion. Classical control concepts of event-based control and perturbation analysis are explored for the system. This work covers modeling the robot's dynamics with simplified assumptions and simulating in MATLAB. Further simulations are conducted in Simscape by importing the robot model. In addition, extensive testing is conducted on the physical system to observe the behavior and performance of trotting gaits.

The main contributions of this MS Thesis is centered around taking steps towards successful multi-modal demonstrations using Northeastern's legged-aerial robot, Husky Carbon. This work
discusses the challenges involved in achieving multi-modal locomotion such as trotting-hovering and thruster-assisted incline walking and reports progress made towards overcoming these challenges. Animals like birds use a combination of legged and aerial mobility, as seen in Chukars’s wing-assisted incline running (WAIR), to achieve multi-modal locomotion. Chukars use forces generated by their flapping wings to manipulate ground contact forces and traverse steep slopes and overhangs. Husky's design takes inspirations from birds such as Chukars. This MS thesis presentation outlines the mechanical and electrical details of Husky 's legged and aerial units. The thesis presents  simulated incline walking using a high-fidelity model of the Husky Carbon over steep slopes of up to 45 degrees.

\end{abstract}

\end{frontmatter}

% --- body of the document ---
\pagestyle{headings}

%% include each chapter like below
% intro.tex:

\chapter{Introduction}
\label{chap:intro}

Bio-inspiration and bio-mimicry offer great promise to the design, development, and capabilities of mobile robots\cite{ramezani2017biomimetic,ramezani2022aerobat,sihite2022bang,sihite2021orientation}. Animals have evolved to be able to traverse across expansive types of environments in nature, with some animals displaying impressive multi-modal capabilities. Researchers have tried to emulate such multi-modal capabilities with the concept of hybrid locomotion mobile robotic systems starting in the early 1980s \cite{ichikawaHybridLocomotionVehicle1983}. What followed was research combining different environmental locomotion capabilities like aerial-wheel \cite{meiriFlyingSTARHybrid2019} and aquatic-wheeled \cite{xingHybridLocomotionEvaluation2018}. Some others include ASGUARD- a leg-wheeled platform \cite{pub4827}, SCAMP- an aerial-climbing platform \cite{popeMultimodalRobotPerching2017}, and HyTAQ - an aerial-terrestrial platform \cite{kalantariDesignExperimentalValidation2013}. 

ANYmal on Wheels is an enhanced iteration of the ANYmal robot, featuring wheels located at the end of its legs\cite{vollenweider2022advanced}, while Max\cite{chi2022linearization}, a research platform created by Tencent, is equipped with wheels positioned at the knee joint. A fascination of researchers has long been the locomotive capabilities of birds. Birds are agile flyers; they can walk rapidly, jump over large obstacles and fly to evade predators. This type of mobility is of high interest in robotic applications for scenarios where only one type of locomotion would be insufficient such as search and rescue missions. Aerial and legged locomotion offers the possibility of long-distance travel with a flight mechanism, while the walking mechanism helps in close inspection and surveillance in terrain where wheeled robots would face difficulty. Research on combining aerial and ground mobility capabilities in a single robot is still in its early stages. In the initial stages, wings were utilized for achieving aerial capabilities with Morphing Micro Air-Land Vehicle (MMALV) \cite{bachmannBiologicallyInspiredMicrovehicle2009} being one of the first. 

Further, such implementations were explored with DALER
\cite{dalerBioinspiredMultimodalFlying2015} with a foldable skeleton mechanism used for both wing and flight. In these designs, efforts to integrate legged and aerial mobility mechanisms in a single platform are often constrained by the divergent design considerations for each system. As such, it can be challenging to fully incorporate the functionalities of both legged and aerial systems in a unified platform. 
% While the bipedal walking and aerial-capable LEONARDO robot represents a successful example of such integration\cite{kim2021bipedal}, the extension of this approach to quadrupedal locomotion and aerial motion remains a relatively unexplored research direction.

\begin{figure}
    \includegraphics[scale=0.6]{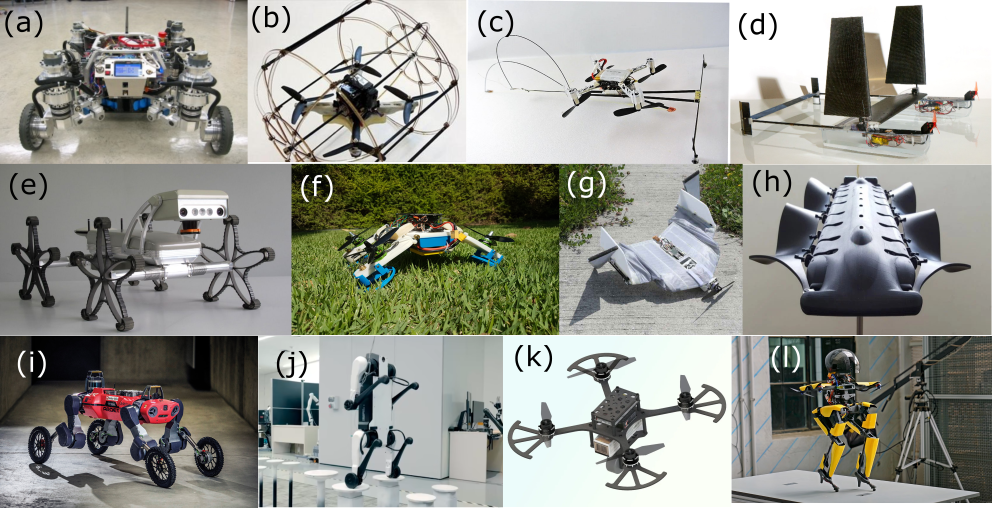}
    \caption{Illustrates state-of-the-art multi-modal robots. (a) Azimuth \cite{bruzzoneReviewArticleLocomotion2012}, (b) HyTAQ \cite{kalantariDesignExperimentalValidation2013} , (c) SCAMP\cite{popeMultimodalRobotPerching2017}, (d) SailMAV \cite{zuffereySailMAVDesignImplementation2019}, (e) Asguard\cite{babuEffectsWheelSynchronization} (f) FSTAR \cite{meiriFlyingSTARHybrid2019} , (g) DALER \cite{dalerBioinspiredMultimodalFlying2015}, (h) Velox \cite{bainesAmphibiousRoboticPropulsive2021}, (i) ANYmal on Wheels \cite{vollenweider2022advanced}, (j) Max \cite{chi2022linearization}, (k) Skywalker \cite{pan2023skywalker}, (l) LEONARDO  \cite{kim2021bipedal}}
    \label{fig:multi-modal-examples}
\end{figure}

Husky Carbon, which is our project inspired by the locomotion of birds, has shown the possibility of integrating quadrupedal-aerial locomotion\cite{salagame2022letter,ramezani2021generative,de2020thruster} into a single platform. This integration is motivated by the need for fast mobility at high altitudes and safe, agile, and efficient mobility in unstructuredspaces\cite{dangol2021control,liang2021rough,sihite2021unilateral,sihite2022efficient}. In certain scenarios, mono-modal mobile systems may easily fail, posing challenges in Search and Rescue (SAR) operations and the aftermath of disasters. For example, Unmanned Aerial Systems (UAS) can perform aerial surveys and reconnaissance, but airborne structural inspection of buildings and mobility inside collapsed structures is challenging. On the other hand, legged mobility is superior in these scenarios. The energy efficiency of legged locomotion\cite{ramezani2014performance,buss2014preliminary} is much higher and as a result operation endurance when legged locomotion is utilized is much higher than flight endurance.

\section{Motivation and Bio-inspiration}
\begin{figure}
    \centering
    \includegraphics[scale=0.3]{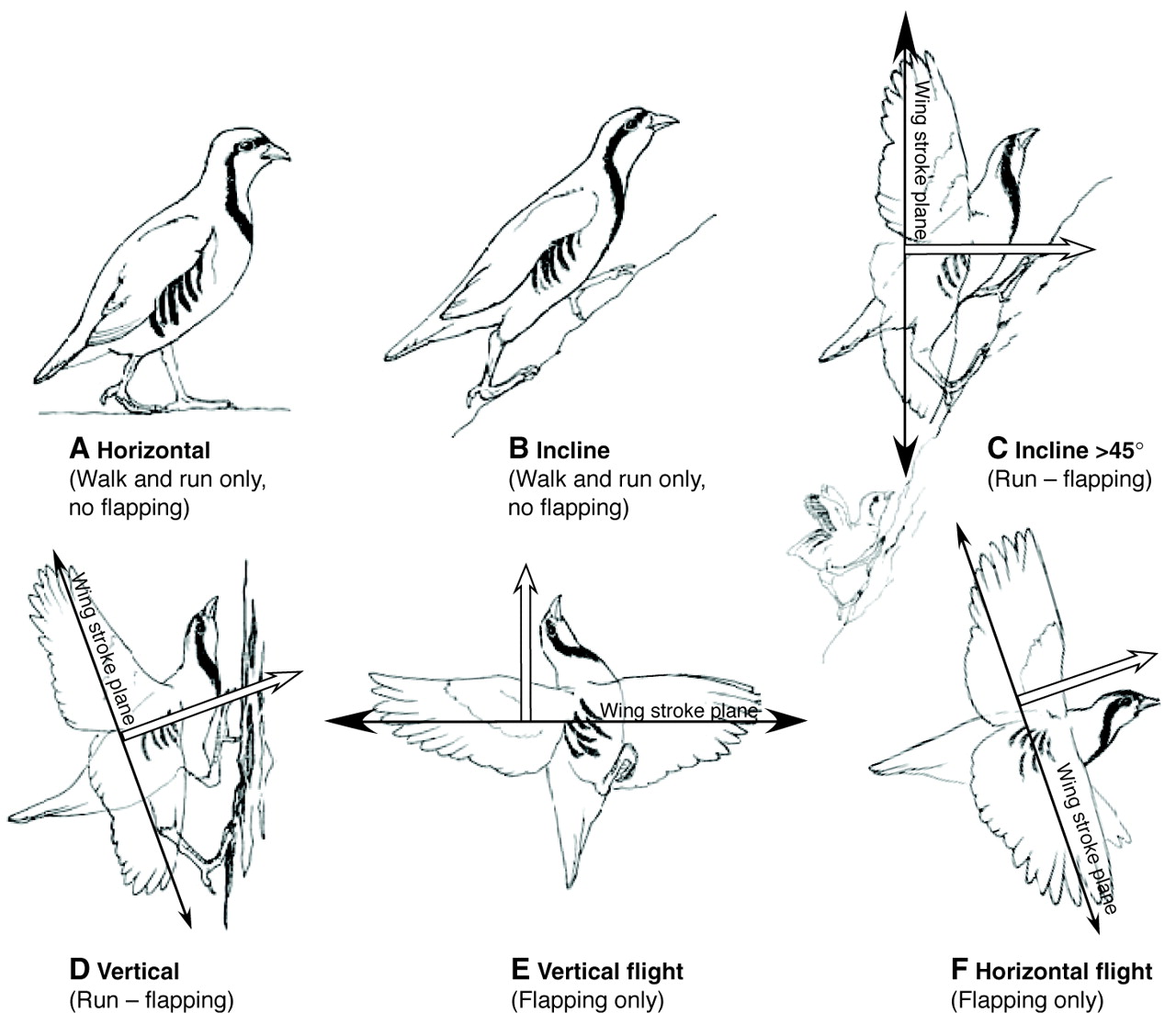}
    \caption{Overview of wing excursions during WAIR and proposed transitions accompanying the WAIR origin of flight hypothesis. (A and B) Birds running over level substrates or shallow inclines do notrecruit their wings to assist running. However, even partial wing development provides assistanceto individuals during incline (45\textdegree) locomotor performance (C). (C and D) A portion of thewingbeat cycle (up to 30\%) involves aerodynamic or inertial forces directed toward the inclinedsurface, rather than skyward, which is sufficient to augment hindlimb traction during WAIR. (D to F) On mastering vertical inclines, birds attain a transverse (dorsoventral) wing excursion that isrequired for aerial flight.\cite{doi:10.1126/science.1078237}}
    \label{fig:bird-WAIR}
\end{figure}
Biomimicry has emerged as a promising approach for designing mobile robots with exceptional capabilities in the field of robotics\cite{sihite2020computational,sihite2021integrated,sihite2020mechanism,ramezani2020towards,ramezani2016bat,hoff2018optimizing,ramezani2017describing,sihite2022unsteady}. By emulating the movement strategies of animals, researchers aim to create robots that can navigate complex terrains with ease. Nature offers a plethora of examples of resilient and fault-tolerant locomotion strategies that animals  use to move around using multi-functional appendages. One such example is the Chukar bird, which employs its wings and legs collaboratively to perform wing-assisted incline running (WAIR)\ref{fig:bird-WAIR}, a highly skilled form of locomotion that enables the bird to navigate steep slopes \cite{hoff2019trajectory, 10.1242/jeb.001701, dial2003evolution}. The ability to perform WAIR expands the Chukar's habitat range, allowing it to interact with and navigate various environments.

Despite the potential benefits of biomimicry, replicating legged and aerial locomotion poses significant challenges for robot design. Achieving the balance required for ground and aerial locomotion is a complex task. Therefore, emulating the motion of WAIR using a robotic platform remains a challenging endeavor. Legged-aerial systems face conflicting design challenges which demote them. For instance, legged robots that can dynamically interact with their environment rely on high-torque actuators which are often bulky. On the other hand, aerial locomotion can be negatively get affected by added mass from these actuators. 

\section{Objectives and Outline of Thesis}
\label{sec:objectives}
The main objective of this thesis project is to:
\begin{itemize}
    \item Design mock hardware of Northeastern Husky Carbon called Husky-$\beta$ to test fliability and legged locomotion of Husky Carbon using a less expensive system to mitigate the risk of failures in preliminary tests 
    \item Generate a high-fidelity model of Husky Carbon in Matlab SimScape to study various maneuvers including legged-aerial and WAIR 
    \item Design and prototype the propulsion unit of the Husky Carbon platform
\end{itemize}

Legged and aerial mobilities are two completely different forms of locomotion that dictate their own requirements. For instance, in the bird example, the bird needs a light body structure to be able to fly. On the other hand, it requires a powerful musculostructural system to support legged locomotion which can be bulky and heavy. A robot design that can accommodate these antagonistic requirements can be hard to control because, for instance, light structures employed to reduce payload for flight introduce compliance that renders locomotion control hard to extremely difficult. Therefore, achieving the overarching objectives of this thesis outlined above can be a significant ordeal. 
% The overarching motivation of this thesis stems from the merits of aerial and legged systems and incorporates the mobility of both systems together into a single platform. Such multi-modal systems offer a great potential of application allowing for flight over long distance and terrain traversal where flight is not possible. Encompassing both capabilities within one system brings several challenges and considerations for the development of stable walking gaits. This thesis introduces and investigated the legged locomotion capabilities of a multi-modal quadrupedal and aerial platform; Husky Beta and Husky Carbon. 

The structure of this thesis is as follows: Chapter 2 introduces the Husky Carbon platform by discussing its design, components, and control difficulties. Additionally, it provides an outline of connected design concepts and control design approaches. Chapter 3 presents the simulation outcomes and system design of the Husky-Beta. The reduced-order dynamic modeling{\cite{dangol2021reduced}} of the system is covered in Chapter 4. Chapter 5 describes the proposed control design method and its implementation using MATLAB. It also covers the Simscape simulation setup and results. Chapter 6 details the physical testing process and the results obtained.

\section{Contributions}
\label{sec:contribution}
This thesis contributes to ongoing research on the design and control of Husky Carbon at Northeastern's SiliconSynapse Lab as follows.

In the Husky Beta project, the hardware design and construction of all aspects of the platform, implementation of leg-end position-based control functionality to achieve a rudimentary trotting-in-place motion, and the development of a high-fidelity MATLAB SimScape Model to demonstrate locomotion and morphing capabilities of the robot in simulation. 

And in the Husky Carbon project, contributions included the development, integration, and testing of the carbon-fiber-composite-based flight system dubbed the Propulsion Unit (PU), expanding the previously developed MATLAB SimScape simulation model to include thruster-assisted locomotion, and conducting rigorous hardware testing of the legged locomotion controller using the real-time kernel of MATLAB Simulink. Finally, this thesis also contributes by designing a servo drive system and integrating it into the customized actuator system of Husky.

%file:///C:/Users/Silicon%20Synapse/OneDrive%20-%20Northeastern%20University/Desktop/Thesis/Legged%20locomotion.pdf
%file:///C:/Users/Silicon%20Synapse/OneDrive%20-%20Northeastern%20University/Desktop/FULLTEXT01.pdf
%file:///C:/Users/Silicon%20Synapse/OneDrive%20-%20Northeastern%20University/Desktop/the_intro_reference_2.pdf
%file:///C:/Users/Silicon%20Synapse/OneDrive%20-%20Northeastern%20University/Desktop/the_intro_reference.pdf

%% husky platform
\chapter{Northeastern's Husky Carbon}
\label{chap:husky}

\begin{figure}[h!]
    \centering    
    \includegraphics[scale=0.37]{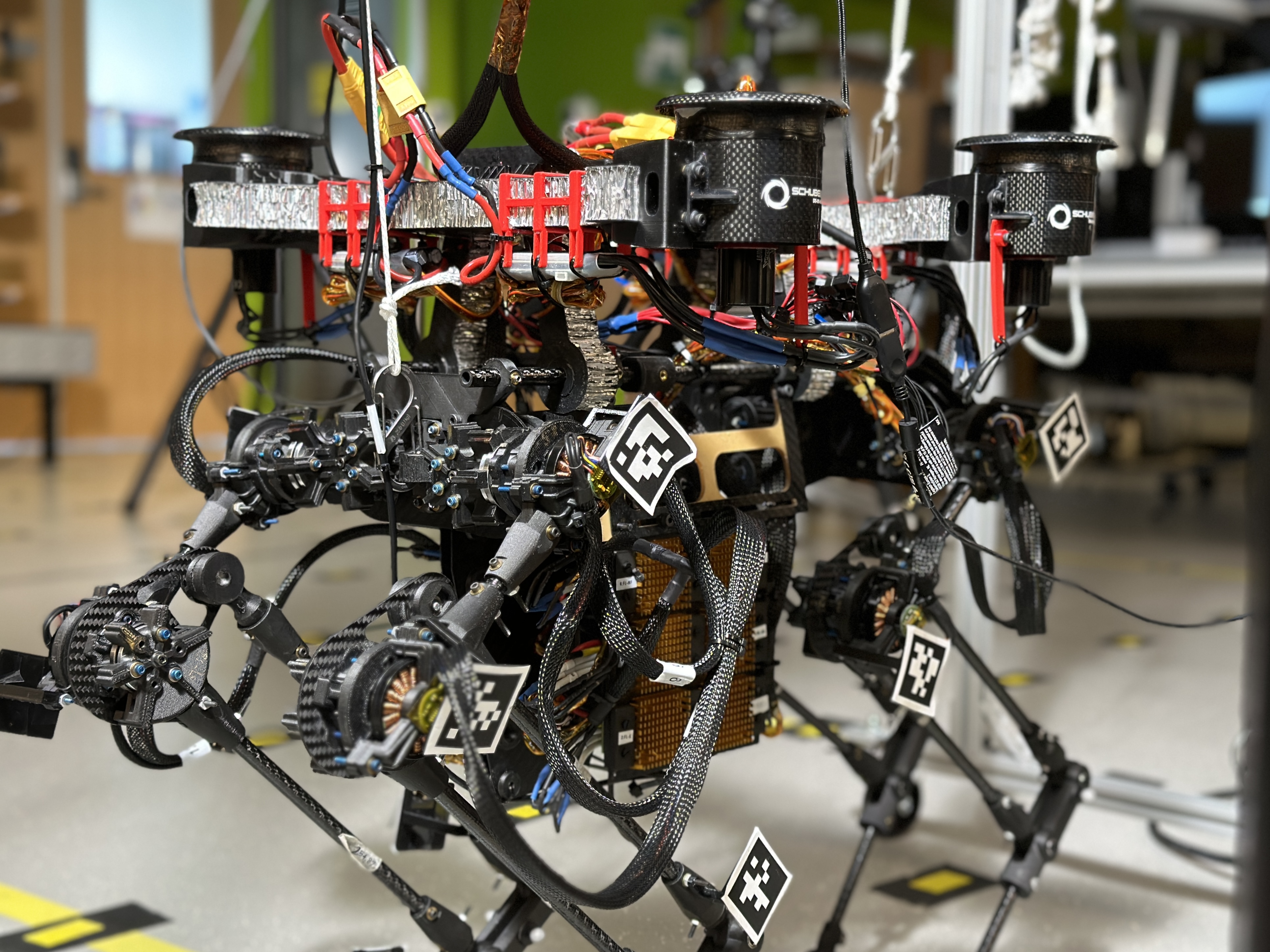}
    \caption{Illustrates Northeastern University’s Husky Carbon Platform, which is a multi-modal legged-aerial robot}
    \label{fig:husky}
\end{figure}
This chapter provides a brief overview of Northeastern's Husky Carbon platform. While Husky's design and prototyping has been the result of efforts from several PhD and MS students, this brief overview is provided here to help the completeness of this Thesis. An overview of the hardware and control design for the legged subsystem is presented in this Chapter.   
\newpage

\section{Northeastern's Husky Carbon: A Multi-modal Legged-Aerial Robot}

% \begin{figure}[h!]
%     \centering    
%     \includegraphics[scale=0.09,\linewidth=1.0]{fig/Husky_2_noBG.png}
%     \caption{\textcolor{red}{Illustrates Northeastern University’s Husky Carbon Platform, which is a multi-modal legged-aerial robot}}
%     \label{fig:husky}
% \end{figure}

\begin{figure}[ht]
    \centering
    \includegraphics[scale=0.3]{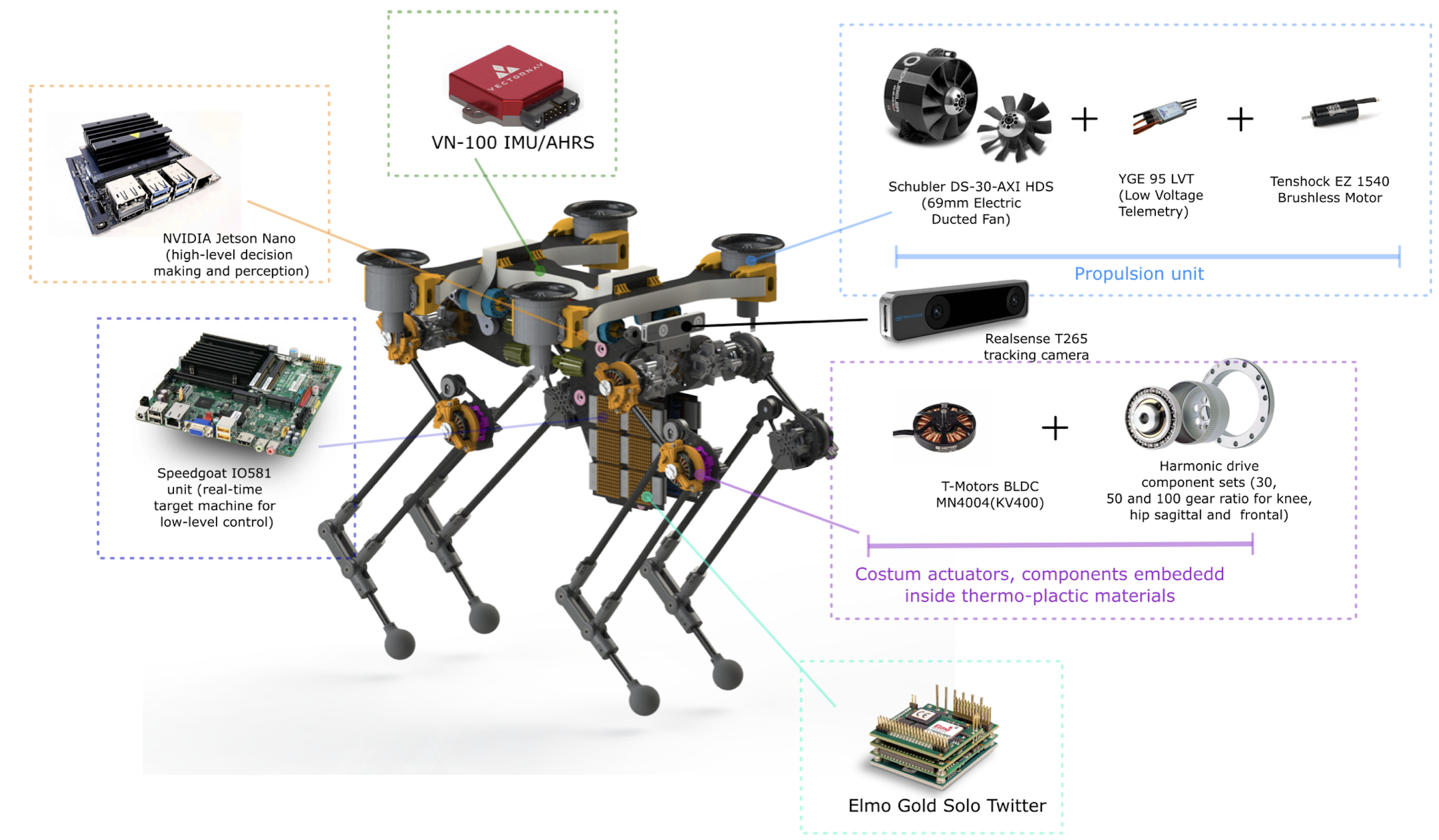}
    \caption{Brief Overview of Husky Carbon}
    \label{fig:husky-overview}
\end{figure}
 Husky Carbon, developed at \textit{SiliconSynapse}, shown in Fig.~\ref{fig:husky-overview}, is a custom quadrupedal robot standing at 2.5ft tall and 12 inches wide. Full leg dimensions can be seen in Table ~\ref{table:leglengths}. It has been fabricated from reinforced thermoplastic materials through additive manufacturing and has a total weight of 7.6kg. The robot is constructed of two pairs of identical legs in the form of a parallelogram. Each leg has three degrees of freedom provided through the Hip frontal (HF) joint, Hip Sagittal (HS) and Knee (K) joints. 
 
 The motor for each joint is the T-motor Antigravity 4006 brushless motor coupled with Harmonic Drive units(CSF-11-30-2A-R). In order to control the amount of current sent ELMO gold Twitter servo drivers are used. A complete list of components is listed in Table ~\ref{table:components}. To facilitate connections with the motor drives, an extension PCB is mounted onto each motor drive providing EtherCAT input/output, safe torque off (STO), and encoder input (RLS RMB20) for joint position feedback. 
 
 The ELMO drives have been isolated from the legs and mounted on a rack on the robot's torso to minimize leg inertia for rapid footplacements. The on-board sensors are a VectorNav VN-100 IMU, Intel Realsense T265, and Hall-effect encoders on each motor. Finally, the propulsion unit uses DS-30-AXI HDS electric ducted fans with the appropriate brushless motor and ESC combination.
\begin{table}[H]
\begin{center} 
\begin{tabular}{cc}
\toprule
 \textbf{Links}        & \textbf{Length}   \\
\midrule  
    Ankle       &  0.17m  \\
\midrule
 Thigh  &  0.15m   \\
\midrule
Calf  & 0.32m \\
\midrule
Hip  & 0.1m \\
\bottomrule
\end{tabular}
\caption{Leg Dimensions in Northeastern's Husky Carbon}
\label{table:leglengths}
\end{center}
\end{table}

\section{Communication Between Modules}
\begin{figure}[h]
    \centering
    \includegraphics[width=1.0\linewidth]{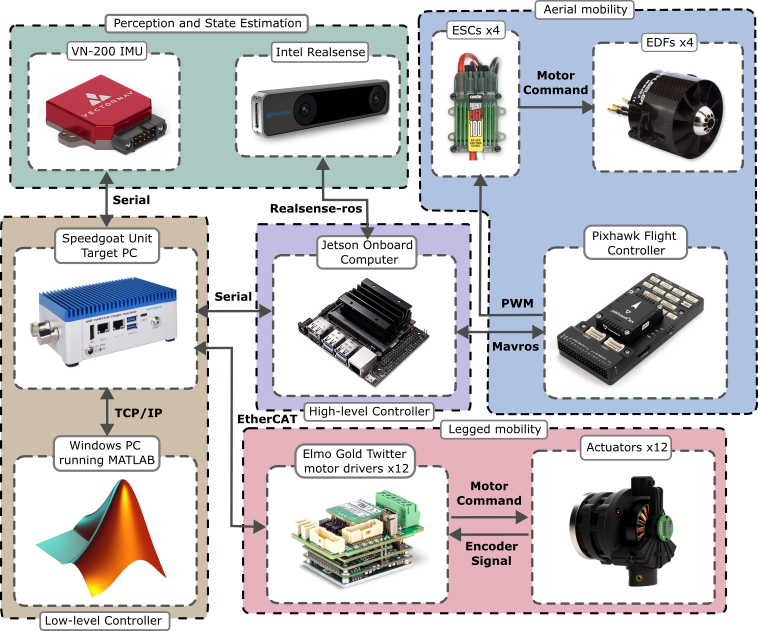}
    \caption{Communication Overview in Northeastern's Husky Carbon}
    \label{fig:comm-overview}
\end{figure}
Figure \ref{fig:comm-overview} depicts the main sub-modules and their communication. The system comprises two major controllers: a low-level and a high-level controller. The low-level controller runs at a rate of 500Hz on a real-time-target processor from Speedgoat, a custom computer with serial and EtherCAT communication capabilities through four RS232 ports and EtherCAT compatible chipset. 

A VectorNav VN-200 IMU provides orientation feedback to the low-level controller via serial to facilitate self-stabilization. The high-level controller runs ROS (Robot Operating System) on an NVidia Jetson Nano and uses an Intel Realsense stereo-camera along with IMU data for path planning and navigation. In addition, Jetson is also interfaced with the Pixhawk Flight Controller Unit (FCU) using the MAVLink protocol for thruster-assisted stabilization and control. 

Human operators interact with the high-level controller over Wi-Fi and the low-level controller over ethernet through a host PC. The target unit communicates with the motor drives via EtherCAT to ensure fast update times and precise synchronization. The Master (Speedgoat) issues position commands to 12 ELMO motor drives at a rate of 10kHz. The motor driver converts the input signal into PWM signals, which control the current transmitted to each of the three phases of the brushless motors. The drives also output signals from magnetic incremental encoders at a rate of 500-4kHz (depending on the maneuvers considered).

% The main sub-modules and their communication can be seen in figure \ref{fig:comm-overview}. The processor for the robot is a unit real-time target machine from Speedgoat with an IO581 unit allowing for serial communication with four RS232 ports. The serial communication interfaces include VN-200 IMU mounted to the body of the Husky sending data at 800Hz and an Intel Realsense T265 camera relaying pose messages at 200Hz. These allow access to body position and orientation values. 
% The Host PC is utilized to create the Simulink Realtime code to be compiled and deployed on the Speedgoat machine. The host PC is connect via EtherNET and allows updates of block parameters within MATLAB while the target is running to control the trajectories and commands sent to the robot. The compiled code is set to run at 500 Hz. 
% The target unit communicates to the motor drives via EtherCAT to allow for fast update time and precise synchronization. The Master (Speedgoat) sends position commands to 12 of the ELMO motor drive at a rate of 10kHz. The motor driver converts the input signal into PWM signals that control the current being sent into each of the three phases of the brushless motors. The drives output back signals from magnetic incremental encoders at a rate of 10kHz as well. 
% \begin{comment}
% \begin{figure}
%     \centering
%     \includegraphics{example-image-a}
%     \caption{Illustrates the embedding process employed to reduce the total payload in the system.}
%     \label{fig:embeding}
% \end{figure}
% \end{comment}

\begin{table}[ht]
\begin{center}
\begin{tabular}{cc}
\toprule
  \textbf{Component}        & \textbf{Name}   \\
\midrule  
    IMU       &  VN-200  \\
\midrule
 Stero Camera  &  Intel Realsense T265  \\
\midrule
Actuators  & T-motor Antigravity 4006 brushless \\
\midrule
Harmonic Drives & CSF-11-30-2A-R \\
\midrule
Servo Drives  &  ELMO gold twitter  \\
\midrule
Encoders  &  RLS RMB20 \\
\midrule
Flight controller  &  Pixhawk 2.4.8\\
\midrule
Electric Ducted Fan (EDF)  &  DS-30-AXI HDS (69mm) \\
\midrule
Electric Speed Controller (ESC)  &  YGE 95A LV \\
\bottomrule
\end{tabular}
\caption{Components on Husky Carbon}
\label{table:components}
\end{center}
\end{table}

\section{Control}
In this section, first, we discuss the control challenges in the Husky platform. Then, we present control ideas that are available and can be utilized. We remain focused on legged locomotion control in this section. Aerial locomotion control does not pose significant challenges since aerial robot control is extensively studied. We present modeling and control design concerning the WAIR maneuver in Sections 5.

\subsection{Legged Locomotion Control Challenges}

The novelty of Husky lies in its ability to achieve multi-modal locomotion, specifically combining legged and aerial mobility with dominant inertial effects. This biomimicry of animals like birds can yield ultra-flexible systems in terms of their ability to negotiate their task spaces. In order to realize such a system, it was designed with strict prohibitive design considerations to allow for flight. This included considerations for limited payload, power budgets, and non-bulky actuators/metal parts which are typically widely used in legged robotics. Thus, the inherent stability often found with quadrupeds is minimized which brings forth new considerations and challenges for gait design and control. 

\begin{itemize}
\item Narrow support polygon - Husky has a low cross-section torso, the hips are located very close to each other in the frontal plane. This is an important design consideration for the quadcopter configuration of Husky in order to to reduce induced drag forces for aerial mobility. It does however present a drawback for the legged mobility of the system. It reduces the support polygon (the convex hull formed by the
ground contact points) leading to a smaller stability margin.

\item Compliance:- Another important consideration for aerial mobility is the total mass of the system. The legs were designed to be as lightweight as possible, avoiding metal and other dense components where possible. Carbon fiber tubes were used for the leg links to minimize the weight of the links. Composite 3D printing techniques were employed for the actuator housing. Two carbon fiber plates are utilized to connect the lower and upper body with the frontal hip harmonic drive housings and have two pelvis plates embedded within them. These approaches, though minimizing weight, reduce the rigidity of the system. Husky suffers from compliance due to these structure bending especially when under load. This increases the challenge of stability, modeling of the system, and anticipated behavior.
\end{itemize}

\subsection{Possible Gaits and Control Design Paradigms}

Here, we briefly explain the control design concept for trotting based on an event-based method developed by previous students and used as a backbone in experiments performed in this Thesis. Later in the thesis, we expand this method with a collocation technique allowing Husky locomotion over steep slopes, i.e., WAIR maneuver.

The overall control design and approach for legged locomotion is dependent on the desired gait to be achieved by the system. Quadrupeds can be set to execute a variation of gaits based on desired overall speed, sequences and strides. The main idea is to contemplate trotting in form of a robot configuration with two contact points with its environment and adjust foot placement based on state feedback to maintain a stable periodic orbit \cite{xuGaitAnalysisQuadruped2019}.

The most common type of gaits for quadrupeds includes crawling (or creeping) and trotting. Within the crawling gait, each leg is raised one at a time sequentially resulting in walking motion. The gait beings with all legs on the ground, in the stance phase. Next, one leg is lifted off the ground entering its swing phase and comes back down instantiating the switching phase at which point it is back to stance and the next leg enters its swing phase. Through this, walking can be achieved in 4 switching periods and 8 total phases of each leg \cite{xuGaitAnalysisQuadruped2019}. Though this produces a rather slow walking gait it requires less energy and is more stable than other gait cycles. 

On the other hand, in the trotting gait, two opposing legs are lifted off the ground at the same time. This involves two legs being in the swing phase and two in stance, making four periods in the gait within one switching phase. A trotting gait allows faster methods to cover distances \cite{xuGaitAnalysisQuadruped2019}.

Other types of gaits include bounding gaits which are generally defined as motion in which the forelegs and the hind legs emphasize with almost the same phase\cite{yamamotoGeneralizationMovementsQuadruped2020}. In this type of gait, support is alternated between pairs of legs, much like the trotting gait. The distinction from the trotting gait is that the fore legs hind legs act in unison instead of the opposing legs. This motion thrusts the body forward allowing for forward propulsion and making it well suited for obstacle avoidance\cite{yamamotoGeneralizationMovementsQuadruped2020}. Bounding gaits generally appear at high speeds and are mainly used by a few small quadrupeds such as squirrels and dogs. A contrast gait from trotting and bounding is a pronking gait in which all legs operate in the same phase\cite{yamamotoGeneralizationMovementsQuadruped2020}. This gait involves all legs being lifted off the ground and touching back down simultaneously. There are only two phases in this gait, in-air, and on-ground. The forelegs generate a thrusting force and the hind legs contribute breaking forces to the motion. Pronking is commonly seen in quadrupedal animals like gazelles and kangaroos. 

The inherent dynamic behavior of a two-contact-point gaitcycle is preferred over the quasi-static behavior of a crawling motion. On the other hand, the two-contact-point gaitcycle is preferred over galloping due to its simplicity. Therefore, in our trotting control, we will consider switching two-contact-point configurations and meanwhile attempt to ensure the stability of periodic orbits. But, this requires us to briefly expand the notion of stability in this context.

\subsubsection{Stability}

For various quadrupedal locomotion gaits, the expected stability achievable is different and mainly attributed to the number and geometry of contact points on the ground and the center of mass (CoM) or center of pressure (CoP) of the robot. A statically stable gait is one in which no feedback or posture control is required at any point to prevent the robot from fallover. More specifically, it is when the CoM of the robot lies within the support polygon which is the convex hull formed by the ground contact points \cite{bottcherPrinciplesRobotLocomotion}. 

The crawling gait for quadrupedal locomotion would be a statically stable gait as long as the CoM lies with the triangle formed by the three leg contacts. Most statically stable walking legged systems are hexapods as they can sustain more contact on the ground throughout the gait, while with quadrupeds, though crawling is stable it results in much slower motions. When the support polygon shrinks in its size, static stability becomes more difficult. For instance, with bipeds static stability is much more difficult to achieve as the support polygon is closer to line and reduced to a point during the single support phase of a gait. Thus, most bipeds have dynamically stable gaits, defined by the bipeds CoP being on the boundary of the support polygon for at least a part of the gait cycle \cite{westerveltFeedbackControlDynamic2018}%[westervelt]
. A two-contact point trotting gait for Husky would also be dynamically stable as when there are only two legs on the ground the support polygon is reduced to a line. Note that other gaits such as pacing and galloping fall along the same lines as the trotting gait in regard to stability.

\subsubsection{Force Closure Condition}

Stability is affected by other factors such as friction cone conditions. Meaning, for a gait to be feasible and to avoid slippage of the legged system, appropriate ground reaction forces and constraints must be taken into consideration. In order to maintain feasible ground contact, the vertical component of the ground reaction force must be non-negative and the ratio of the horizontal component to the normal component must not exceed the coefficient of static friction given by

 \begin{equation}
 \begin{aligned}
 \frac{Fh}{Fz} &< \mu\\
 Fz&> 0\\
 Fh &= \sqrt{Fx^2+Fy^2}
 \end{aligned}
 \end{equation}
 
\noindent where $F_h$, and $F_z$  denote the horizontal static friction force composed of $F_x$, $F_y$  and the vertical normal force respectively. This set of equations constitutes the friction cone with which the normal force must reside to ensure the equilibrium of the contact (i.e., force closure conditions \cite{focchiSlipDetectionRecovery2018}). For a more simplified instance, it is often common to linearize the friction cone and estimate it as a friction pyramid \cite{focchiSlipDetectionRecovery2018}. These feasibility conditions are often considered in the context of nonlinear-optimization control of legged systems. 

In our control design concept for Husky, we consider design-based Poincare return maps which permit closed-loop adjustment of the two-contact-point trotting gaits without any optimization\cite{sihite2021optimization}. Hence, our approach can yield inexpensive computations, which can be easily executed in real time.

\subsubsection{Poincare Return Map}

Continuing on the notion of stability for the two-point-contact trotting gait, we focus our attention on forming periodic orbits in the underactuated dynamics of Husky. Note that, unlike the general assumption of overactivation for quadrupedal robots, Husky at its two-contact-point trotting with its point contacts is underactuated. Therefore, closed-loop motions, such as walking, trotting, galloping, and running generate periodic orbits i.e limit cycles.

Dynamically stable walking corresponds to the existence of these limit cycles in the robot's state space \cite{westerveltFeedbackControlDynamic2018}
. In order to determine the existence and stability of limit cycles, Poincare sections and return maps are widely used. Briefly speaking, the method of Poincare involves sampling the solution of a system according to a defined rule and evaluating the stability properties of equilibrium points (or fixed points). The existence of these fixed points corresponds to periodic orbits. More concisely, Poincare return maps transform the problem of finding periodic orbits into finding fixed points on a map. 

The Poincare method remains an important tool in determining the stability for hybrid systems consisting of several time-invariant ordinary differential equations linked by event-based switching mechanisms as well as for other broad ranges of system models.  Therefore, the intuition behind our event-based control design paradigm for Husky's two-contact trotting gait is strategically readjusted foot placement trajectories such that a fixed point is achieved.

\subsection{Literature Review on Control Methods}

Other control design approaches widely used in quadrupedal locomotion are listed below:

% \begin{itemize}
%     \item Passive dynamic walking
%     \item Zero moment point
%     \item Foot placement; posture control
%     \item Hybrid zero dynamics
%     \item Reinforcement Learning tecnhinque
%     \item Whole body control
%     \item Reduced order models; slip model; hybrid models
%     \item Model predictive control
%     \item transerve linearization
%     \item event-based controllers
%     \item Central pattern generators (CPGs)
%     \item Explicit reference governors
%     \item Gait regulations
%     \item optimal control
%     \item Approximate Dynamic Programming
%     \item Gait optimization
%     \item Sequential Linear Quadratic
%     \item iLQR, iLQG
% \end{itemize}

% \label{chap:intro:lit review}
\subsubsection{Zero Moment Point}
The Zero Moment Point (ZMP) was the turning point of bipedal locomotion and set the benchmark for future control techniques \cite{vukobratovicZEROMOMENTPOINTTHIRTY2004}. Introduced by Vuckobratoic, the zero moment point is defined as the point where all the forces generated by the contact of the leg end and the ground can be replaced by a single linear force, hence no moments produced. This control technique relies on ensuring the ZMP is within the support polygon made by contact feet through the gait cycle to prevent rolling and maintain stability \cite{vukobratovicZEROMOMENTPOINTTHIRTY2004}.
The ZMP requires the feet to be in contact with the ground and within the support polygon created which causes difficulty in achieving some type of gaits that are more efficient and faster in quadrupedal locomotion. When the support polygon is reduced to a line or a point it is more of a challenge to ensure the ZMP falls within it and it will be unable to achieve the given balance condition. This is especially true for bipeds which require flat feet for ZMP implementation, when the environment becomes unstructured or slanted outputting a desired gait with only ZMP design will cause the robot to fall over. Therefore robots utilizing ZMP are often coupled with another control implementation as well. Such is the case of the popular example Asimo by Honda. It uses ZMP controller coupled ith foot landing position control and posture control \cite{shigemiASIMOHumanoidRobot2019} . Many more implementations have been conducted using ZMP for bipedal locomotion \cite{yamaguchiDevelopmentBipedalHumanoid1999, kajitaZMPBasedBipedRunning2007, sardainForcesActingBiped2004, hun-oklimControlRealizeHumanlike2000,dangol2021hzd} as well as with extensions and adjustments added to it such as preview control theory \cite{kajitaBipedWalkingPattern2003} or allowing variations of ZMP dependent on walking pattern \cite{bum-jooleeModifiableWalkingPattern2008}. The ZMP method is most popular for bipedal locomotion but has been implemented for quadrupedal locomotion as well \cite{liGaitPlanningStability2016, akbasZeroMomentPoint2012,winklerFastTrajectoryOptimization2017}. The ZMP control method has had great contribution to the development of legged locomotion and provides an instinctive understanding of the control necessary but it is not an all encompassing solution to maintain the stability of the system due to it's reliance on equal footings and reduced order models. Hence, ZMP is often coupled with other controller to offer corrective measures whenever there is a breach in ZMP stability criterion.  

\subsubsection{Hybrid Zero Dynamics}
Hybrid systems can be described as a mixture of continuous dynamics and discrete events \cite{westerveltFeedbackControlDynamic2018}. Hybrid modeling is  a popular and versatile technique to describe  the dynamics of a legged system as it allows exchanging models to account for modified physical constraints of the system (such as change number of contact points, i.e impact events).  In this case, stability is often defined with respect to periodic solutions of the system. This is commonly investigated using Poincare map analysis at certain phases of the hybrid system \cite{westerveltFeedbackControlDynamic2018}. 
A well-liked approach is the extension of the ZMP to hybrid modeled legged systems, coined Hybrid zero dynamics (HZD).  
Hybrid zero dynamics allow for a lower-dimensional model of bipedal systems that is still capable of describing the essential features of the locomotion \cite{westerveltFeedbackControlDynamic2018}.  It relies on driving a set of virtual constraints to zero using feedback control. These virtual constraints map the error of the states in a desired trajectory to an output defined in a subspace noted as the zero dynamics manifold \cite{westerveltFeedbackControlDynamic2018}. HZD is mostly common and tailored to bipedal locomotion and has seen successful implementations in simulation and experimentation \cite{fuProvingAsymptoticStability2006b,RABBITTestbedAdvanced2003, hereidDynamicHumanoidLocomotion2018,martinStableRobustHybrid2017}. HZD application to quadrupeds becomes more complex due to the increased degrees of freedom, richer contact forces which  lead to non-trivial design for the virtual constraints and corresponding zero dynamics manifold. Efforts have been conducted to extend the HZD for quadrupeds and mainly rely on utilizing linear and bilinear matrix inequalities \cite{hamedDynamicallyStable3D2019,maFirstStepsFull2019}.

\subsubsection{Model Predictive Control}
Model predictive control (MPC) is a control strategy that has gained popularity in the field of quadrupedal locomotion control. This approach involves using a predictive model of the system dynamics to generate a control sequence that minimizes a cost function, subject to constraints on the system dynamics and control inputs. By incorporating physical constraints such as joint limits and contact forces, MPC can generate dynamically feasible trajectories that ensure the robot remains within its physical limits while maintaining stability\cite{wanner2022model}. One of the key advantages of MPC-based locomotion control is its ability to handle nonlinearities and uncertainties in the system dynamics. Quadrupedal locomotion is a complex process that can be affected by various factors such as friction, terrain, and external disturbances. Traditional control methods may struggle to handle these uncertainties, but MPC can adapt to them and generate control sequences that are robust to changes in the system dynamics. This makes MPC a suitable approach for quadrupedal locomotion control in real-world environments where uncertainties are common.
The effectiveness of MPC-based quadrupedal locomotion control has been demonstrated in several research studies. For instance, \cite{di2018dynamic} used MPC to determine ground reaction forces. The results showed that the robot was able to walk with stable and synamic gait patterns on both flat and uneven terrain. Researchers in Eurpoe used MPC to achieved torque level whole-body Model Predictive Control (MPC)\cite{grandia2019feedback} at the update rates required for complex dynamic systems such as legged robots.
Despite its advantages, there are also limitations and challenges associated with the use of MPC in quadrupedal locomotion control. 
One of the main challenges is the computational complexity of the optimization problem, which can make real-time implementation difficult. Additionally, the accuracy of the predictive model and the constraints used in the optimization problem can affect the performance of the control sequence generated by MPC.
Future research in this area could focus on improving the accuracy and computational efficiency of the predictive model, as well as exploring new methods for generating control sequences using MPC. There is also potential for MPC to be combined with other control strategies such as reinforcement learning to further improve the performance of quadrupedal locomotion control.

\subsubsection{Event-Based Control}
Event-based controllers provide feedback action on outputs sampled based on discrete criteria rather than continuous sampling \cite{aranda-escolasticoEventBasedControlBibliometric2020a}\cite{park2012finite}. It is of note that there is a lack of uniformity to the surrounding the terminology of event-based controllers. In general, they may be distinguished as "event-triggered" i.e based on measurements or "self-triggered" based on predicted dynamics \cite{aranda-escolasticoEventBasedControlBibliometric2020a}. For the purposes of legged locomotion, the "event-triggered" concept is employed. This approach utilizes Poincare sections to design stabilizing feedback controllers for periodic orbits by updating parameters which are held constant over the continuous dynamic phase \cite{westerveltFeedbackControlDynamic2018}. Parameter adjustments take place upon each impact (swing leg touching the ground, when trajectories cross the Poincare section) based on ground reaction forces\cite{westerveltFeedbackControlDynamic2018}. Event-based controllers have been extensively analyzed for bipedal locomotion \cite{hamedEventBasedStabilizationPeriodic2014b, westerveltExperimentalValidationFramework2004c,westerveltSwitchingPiControl2003a} and have been implemented in simulation and practice \cite{wahrmannTimevariableEventbasedWalking2018a, grizzleHZDbasedControlFivelink2008a, ramezaniPerformanceAnalysisFeedback2014a, sreenathEmbeddingActiveForce2013a}. Such an example is work done on ATRIAS 2.1 showing the use of symmetry to simplify and improve design for event-based controllers \cite{hamedEventBasedStabilizationPeriodic2014b}. In comparison, the literature present on using event-based controllers for quadrupedal locomotion is rather lacking, though not absent.Hamed et al. explores the approach within a hierarchical control algorithm using event-based MPC and QP based virtual constraints to generate stabilizing patterns on a quadrupedal robot, Vision 60 \cite{hamedQuadrupedalLocomotionEventBased2020a}. MIT's Cheetah 2 also employs an event-based controller for gait stabilizing of periodic orbits. It was also inspected in experiments on MIT's Cheetah 2 where an impulse-based gait design was used to generate periodic orbits based on a time-switched model and an event-based controller was utilized for gait stabilization \cite{parkHighspeedBoundingMIT2017a}. This work layered continuous body state feedback to manage balance and converge trajectory during stance phase and saw successful experimental results. The merits of event-based controllers lie in their simplicity and computational efficiency, though their inferiority appears from the delay in the discrete event and the augmentation to the dynamics of the system.  

\subsubsection{Reinforcement Learning based control}

Reinforcement learning (RL) has emerged as a powerful approach for controlling quadrupedal locomotion, allowing for the development of control policies that can adapt to changing environments and achieve high levels of performance\cite{jia2022seil}.
One of the key challenges in using RL for quadrupedal locomotion is the high-dimensional state and action spaces involved. This makes it difficult to design effective reward functions and to efficiently explore the space of possible policies. Several recent studies have addressed these challenges by using a variety of RL algorithms and neural network architectures.
One promising approach is to use deep reinforcement learning (DRL) to learn control policies directly from raw sensor data, such as camera images or joint angles, which also knows as model-free RL. This has been demonstrated in several recent studies, such as \cite{margolis2022rapid} \cite{margolis2021learning} \cite{kumar2021rma}, which used a DRL algorithm to learn locomotion policies for a quadruped robot. The policies learned by the algorithm were able to achieve a variety of locomotion tasks, including walking, trotting, and galloping, and were robust to changes in terrain and other environmental factors.
Another approach is to use hierarchical RL to decompose the problem of quadrupedal locomotion into a series of sub-tasks, each of which can be learned independently. The fundamental reason for the importance of a hierarchical policy is the nonlinearity of motor dynamics which is hard to model by using the traditional method. 
One example is the work by ETH Robotics System Lab \cite{hwangbo2019learning}, which used a hierarchical RL algorithm that combined an actuator net and a control policy and then deploy it on a real system. After that, multiple works show the potential of hierarchical RL-based locomotion controller \cite{lee2020learning}, \cite{miki2022learning} \cite{yang2023neural}.
In addition to RL, other machine-learning techniques have also been applied to the problem of quadrupedal locomotion. For example, the reduced order model is always been used for reducing the optimization computation load, but the gap between the reduced-and full-order models always exists, people using machine learning to reduce this gap \cite{pandala2022robust}. Due to the high computation load of traditional optimization, the update frequency of the Model Predictive Controller is limited, people use imitation learning to train a policy that is guided by the solutions from MPC \cite{carius2020mpc}, which remains the same performance but reduced computation need.
Despite these advances, there are still several challenges that need to be addressed in using RL for quadrupedal locomotion. One of the main challenges is the need for large amounts of training data, which can be time-consuming and expensive to collect. Another challenge is the difficulty of ensuring the safety and stability of the learned policies, especially in complex and dynamic environments.

% Some other popular control methods explored in research for quadrupedal locomotion include Reinforcement Learning \cite{shiReinforcementLearningEvolutionary2021, bellegardaRobustHighspeedRunning2021} Model predictive control \cite{dingRealtimeModelPredictive2019, dingRepresentationFreeModelPredictive2021, shiModelPredictiveControl2019}, Whole body control \cite{dariobellicosoDynamicLocomotionWholebody2017, raiolaSimpleEffectiveWholeBody2020, dariobellicosoDynamicLocomotionWholebody2017}, Central pattern generators (CPGs).\cite{huashanfengConstructionCentralPattern2008, rutishauserPassiveCompliantQuadruped2008}.

%% husky platform
\chapter{Husky-$\beta$ Platform}
\label{chap:huskybeta}
\begin{figure}[!ht]
    \centering  
    \includegraphics[scale=0.25]{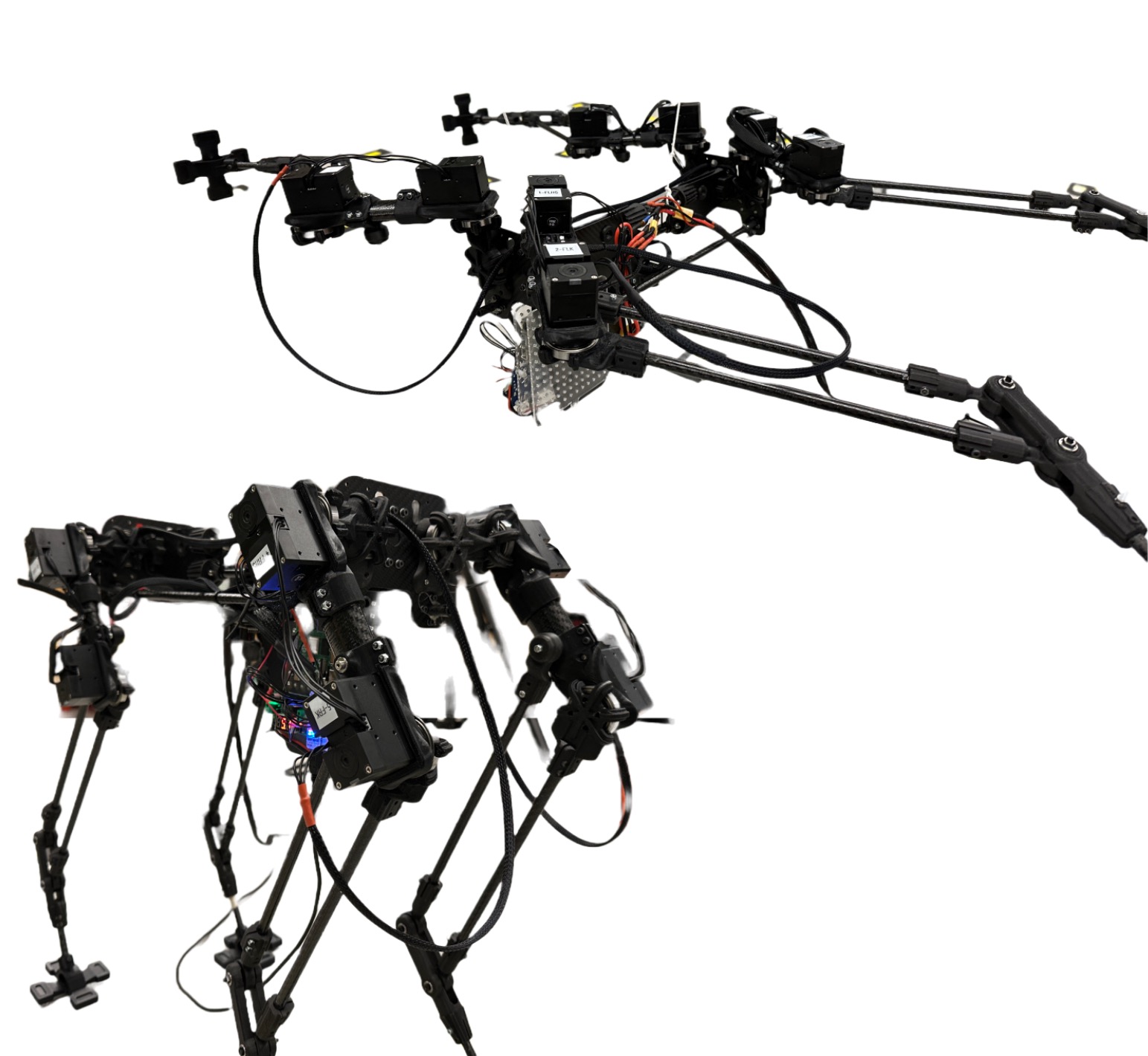}
    \caption{Husky-$\beta$ hardware platform}
    \label{fig:huskyBeta}
\end{figure}
\newpage

\section{SYSTEM DESIGN}

This section discusses the mechanical and prototyping of Husky, which is followed with the details on the electronics and controller architecture. We will also elaborate on the simulation environment used in this work for showing the proof-of-concept.

\subsection{Mechanical Design and Prototyping}
\begin{figure*}[h!]
    \centering
    \includegraphics[scale=0.1]{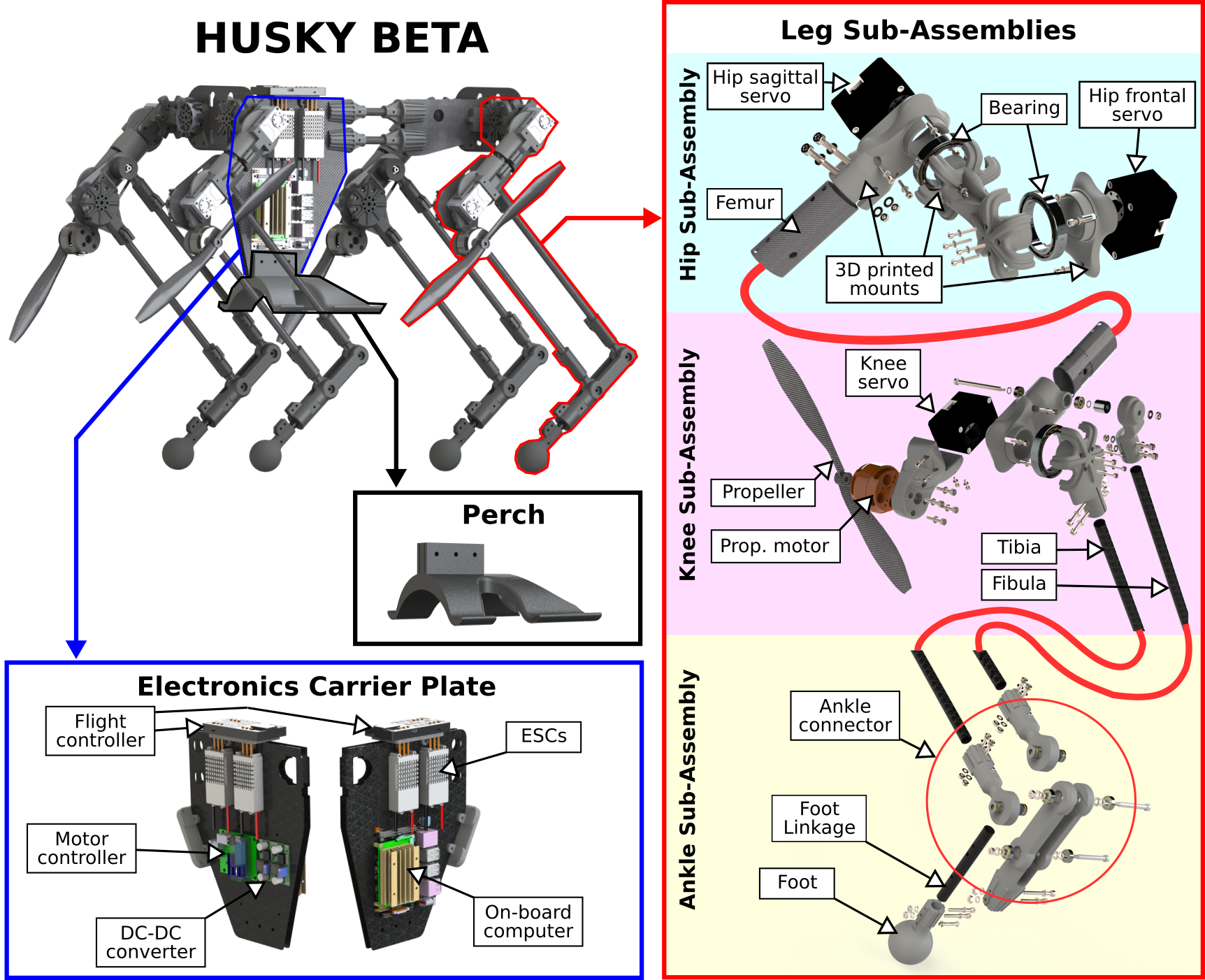}
    \caption{Husky Beta components overview. The system is composed of the main body where the electronics are mounted, and the leg sub-assemblies. The mounting frames and linkages are made out of 3D printed plastic (Markforged Onyx and reinforcing materials) and carbon fiber tubes/plates, respectively. The lower leg, shown as the tibia and fibula linkages, are parallel linkages which is actuated at the knee by the servo. The robot is capable of transforming into the UAV mode and use the propellers mounted at the knee joint for aerial mobility.}
    \centering
    \label{fig:husky_beta_overview}

\end{figure*}

The design of Husky Beta intends to achieve both quadrupedal mobility and multi-rotor flight within the same mechanical architecture. To this end, a propeller motor is attached to the outside of each knee joint, allowing the robot to morph into a quad-rotor configuration by extension of the hip frontal joints. Figure \ref{fig:husky_beta_overview} shows the components of the robot and the sub-systems present in the design. There are three actuated degrees of freedom per leg: hip frontal flexion/extension, hip sagittal flexion/extension, and knee flexion/extension. To simplify the design for this initial prototype, off-shelf servomotors are used to actuate each joint in lieu of lighter, more specialized custom hardware. Extensive use of carbon fiber epoxy laminates fortify the airframe and leg bones, while 3D printed components with carbon fiber reinforcement serve as connecting members. The electronics are mounted on two vertical carbon fiber plates to yield a minimized Total Cost of Transport (TCoT) and payload which was the result generated HuskyCarbon's Mobility Value of Added Mass (MVAM) problem.

%% The morphing process works by first lowering the robot using the knee and the sagittal actuators until the perch touches the ground. 
%% The legs then splay out with the help of the frontal motors which turn roughly 90 degrees. 
%% The servo's can then lock the legs during flight. This can also be done using mechanical locks. This can be advantageous the servos then do not need to constantly out put a torque to lock the leg and consequently the energy efficiency of the robot.

% This work investigates the first of multiple proposed morphing procedures. They are first classified by ”Static" or “Dynamic" based on whether active balancing is required during the transition, and then classified by the number of legs in transition at any one time. These morphing procedures vary in the amount of agility required by the robot in between legged and aerial control, and in the peak thrust required by the propellers. In order of descending difficulty, the authors propose the following:

% In dynamic morphing with all legs, the robot must transition all four legs from legged mode to flight mode simultaneously. This may be achieved by the robot performing a powerful leap into the air and completing all the morphing steps quickly before striking the ground again. This procedure is expected to be highly strenuous on the current supply, the margins of control and stability, and margins of mechanical failure. It may also require a joint torque and propeller thrust largely unattainable by a robot with the architecture from this study.

\begin{figure}[h!]
    \centering
    \includegraphics[scale=0.3]{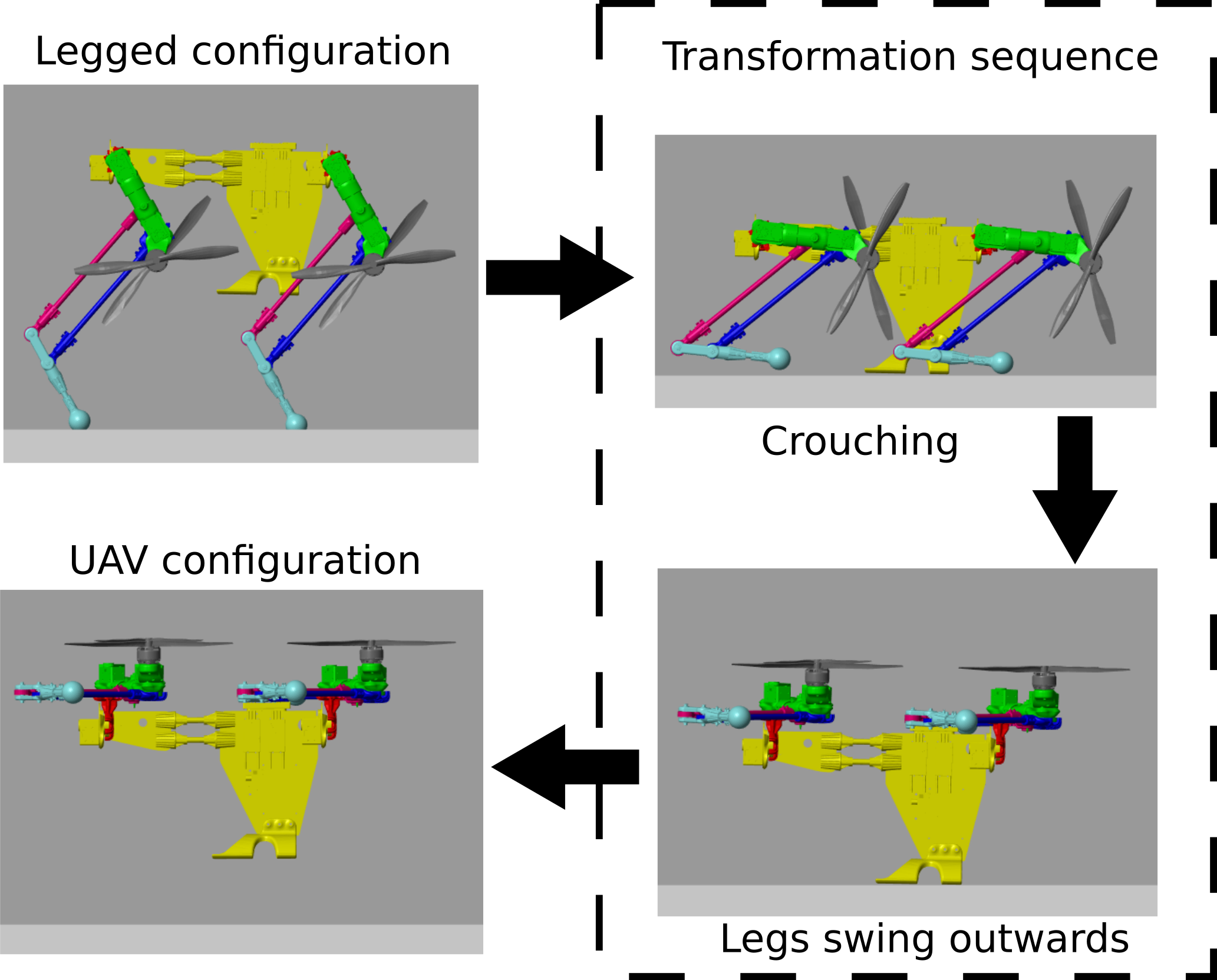}
    \caption{Morphing process simulated in MATLAB Simulink environment using Simscape Multi-body Contact Forces library. The robot transform from its initial legged configuration by crouching, and lifting the leg outwards into the UAV configuration.}
    \label{fig:Morphing}
\end{figure}

The simplest morphing procedure and the one addressed in this work employs a rigid structure fixed to the underside of the robot's body termed a "perch" which shows in \ref{fig:Morphing}. The morphing process starts by the robot lowering the body crouching until the perch comes into contact with the floor. The feet continue to lift, and the weight of the robot is transferred from the feet onto the perch. Once the robot is passively balancing on the perch, the transformation from the crouch position to the aerial position begins. The legs first splay outwards via the frontal servos until the femur and tibia are parallel to the floor. Then, the sagittal servos sweep the propellers horizontally until the center of lift coincides with the robot CoM. At this point, the servo positions remain fixed, and the propellers are powered on. Now the morphing procedure is considered complete, and the robot assumes a standard quadcopter control scheme.

% \begin{enumerate}[i]
%   \item Dynamic Morphing, All Legs
%   \item Dynamic Morphing, Two Legs
%   \item Dynamic Morphing, One Leg
%   \item Static Morphing
% \end{enumerate}

For the reverse (aerial to legged) the robot can make a powered descent using its propellers and land on its perch. Once the full weight has been transferred on to the perch, the propeller motors can turn off, and the legs can then move to the crouch position by undoing the steps listed previously. When the feet come into contact with the floor, the legs push the robot off the perch using the sagittal and knee actuators, and the quadrupedal control scheme is resumed.

An important capability for this system is thrust vectoring. While the flight controller may not benefit significantly from this feature, thrust vectoring becomes an essential mechanism of control during the one-legged and two-legged morphing procedures. During these transitions, the propellers produce imbalanced drag torques in the yaw axis. With the ability to tilt the propellers via small variations in the frontal actuator position, the robot is able to generate corrective yaw torques to remain stable. 

In aerial configuration, the robot locks the joint positions by arming the actuators. While this requires a constant power draw, it is also the simplest solution. A possible improved solution involves a locking mechanism which can mechanically constrain the femur bone relative to the body of the robot. Such a lock can also reduce vibrations by increasing the flexural stiffness of the leg structure, which can prevent unstable oscillations from forming in the flight controller.  The authors are considering the two options for future studies to improve the robustness and resilience of the robot in the aerial configuration. 

\subsection{Electronics Architecture}

\begin{figure}[h!]
    \centering
    \includegraphics[scale=0.55]{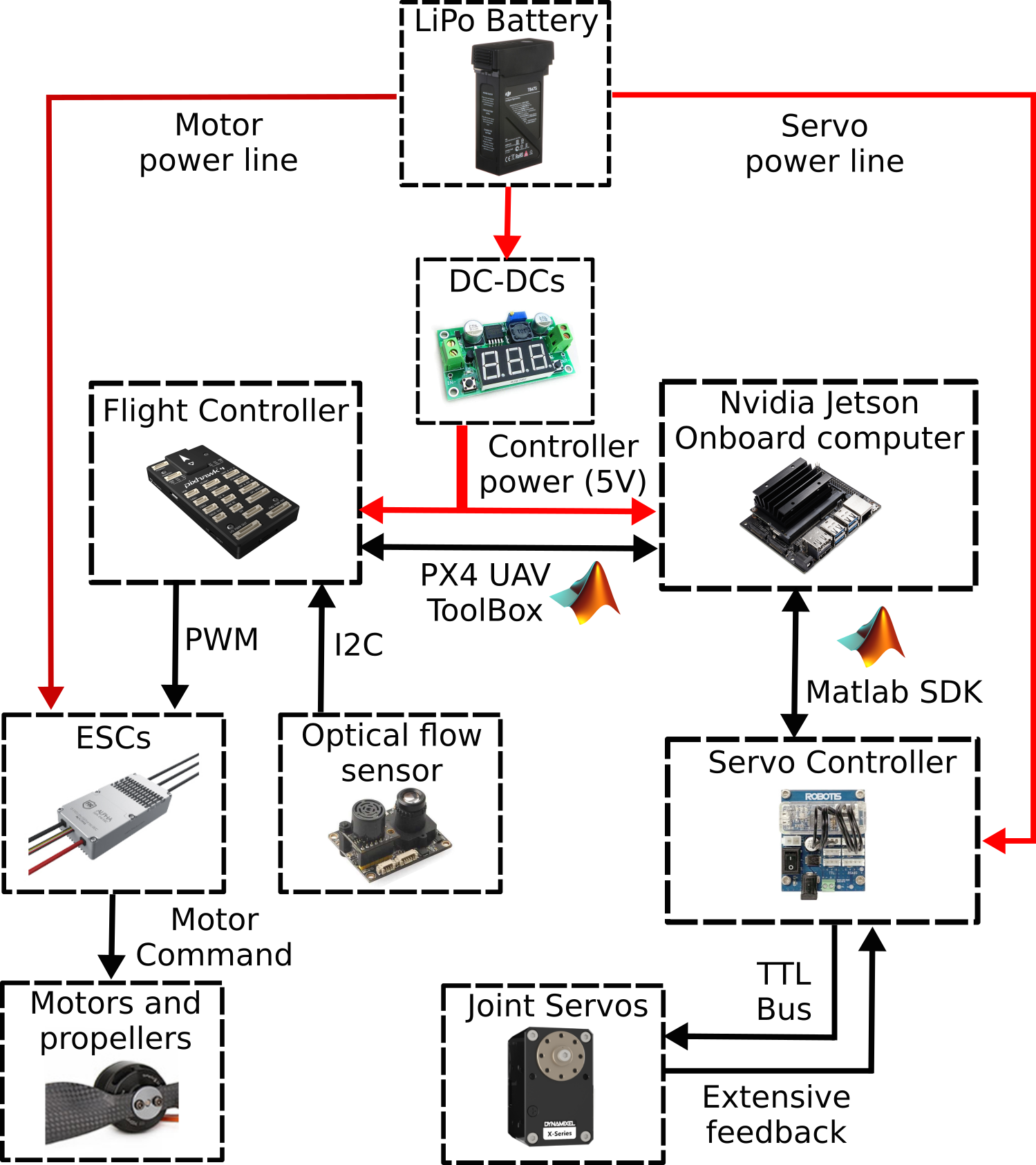}
    \caption{Diagram showing the electronics architecture of the Husky-Beta. A Lithium Polymer (LiPo) battery powers the entire system, which consists of 4 propeller motors and 12 joint servos. The microcontroller units, which consists of the flight controller and an on-board computing unit, coordinates the input/output of the system to interface with the sensors and controller commands and stabilize the robot.}
    \label{fig : electronics architecture}
    \centering
\end{figure}

%% the cite here should be husky carbon ICRA2021
According to the results generated by HuskyCarbon's MVAM problem \cite{ramezani_generative_2021}, all electrical components are mounted on a vertical structure to reduce the TCoT. Figure \ref{fig:husky_beta_overview} shows the electronic carrier plate which forms a vertical structure which carries all essential electronics of the robot. The diagram of the electronic system architecture can be seen in Fig. \ref{fig : electronics architecture}.

This carrier plate follows a sandwich design, consisting of two carbon fiber plates embedded in 3D printed components. The battery base is attached on the lower rear side of the main body structure to ensure that the Center of Mass (CoM) in flight configuration is aligned with the center of lift created by the four propellers. A lithium-polymer battery is selected with a nominal voltage of 14.8 V and a mass of 240g, ensuring an appropriate power to weight ratio. The on-board computer, a Quad-core ARM Cortex-A57 processor, is mounted on one side of the body structure and a servo controller and DC-DC converters are attached to the other side to maintain a balanced lateral mass distribution. Meanwhile, electronic speed controllers are fitted on both sides. On the top, the flight controller which deliver attitude data and propeller motor signals is located.

\newpage

\subsection{Simulation Work}

%%%%%%%%%%%

In order to verify the feasibility of the controller, we first developed the model in in Matlab, and then a simulation is carried out inside the Simscape Multibody physics environment. By designing the appropriate commands for the robot in Simulink and watching the robot's simulated response in Simscape, the joint trajectories, joint torques, and propeller thrusts can be extracted from the results of the simulation. From there, the torque and thrust data are used to justify the actuator selection, and inform further refinement of the gait commands and transformation commands. 

For the controller, the foot position signals that facilitate a forward trot and facilitate the transformation from standing to flying are programmed in Simulink, where foot positions are measured relative to the body coordinate frame of the robot. The trotting gait is based on a ``two-beat gait," sometimes called a ``diagonal gait" where two diagonal legs step forward while the opposing two diagonal legs are planted. An inverse kinematics algorithm resolves the four foot position signals into twelve joint angle signals. In the simulated environment, this controller works by prescribing the positions and velocities of the twelve joints assuming that there is no limit to the amount of torque the joints are capable of providing, and no limit to the amount of thrust the propellers can generate. 

These discrete joint angle signals are transformed into physical signals and passed into the Simscape environment. In the Simscape environment, the robot's solid model is placed on a smooth plane with contact forces modelled between the feet and the floor and between the perch and the floor. The floor used in this study is an elastic ground model with no surface features or obstacles modelled at this time. By selecting both the actual joint trajectories realized in the physics environment and the desired joint trajectories computed by the inverse kinematics algorithm as the outputs of the simulation. %, the difference between the controller's commands on the robot and the robot's resulting motion can be plotted against one another as shown in the 12 subplots of Fig. \ref{fig:Joint Torques during morphing}.  

\begin{figure}[h!]
    \centering
    \includegraphics[scale=0.8]{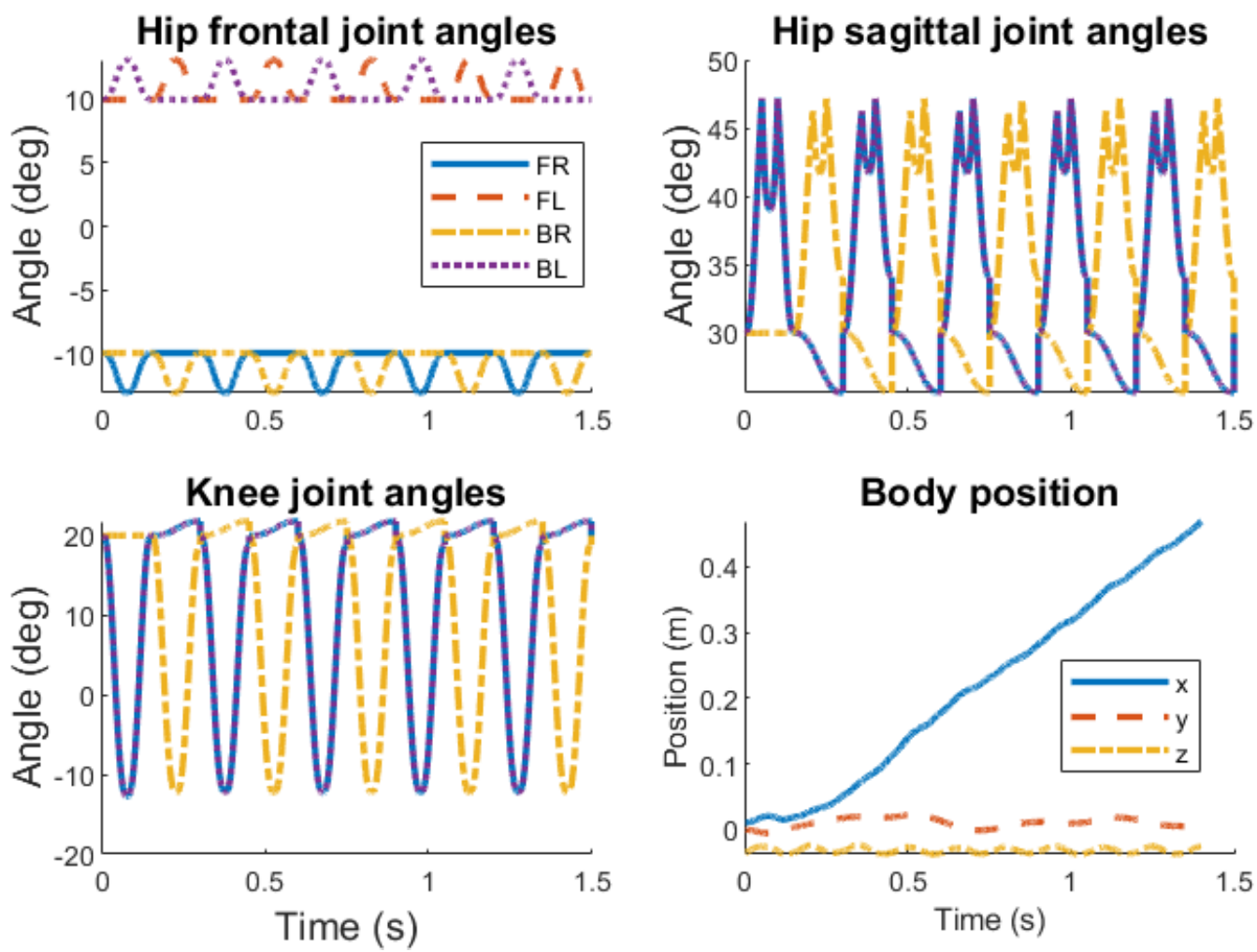}
    \caption{The simulated joint trajectories and body positions during walking using trotting gait. }
    \label{fig: trajectories}
\end{figure}

\begin{figure}[h!]
    \centering
    \includegraphics[scale=0.8]{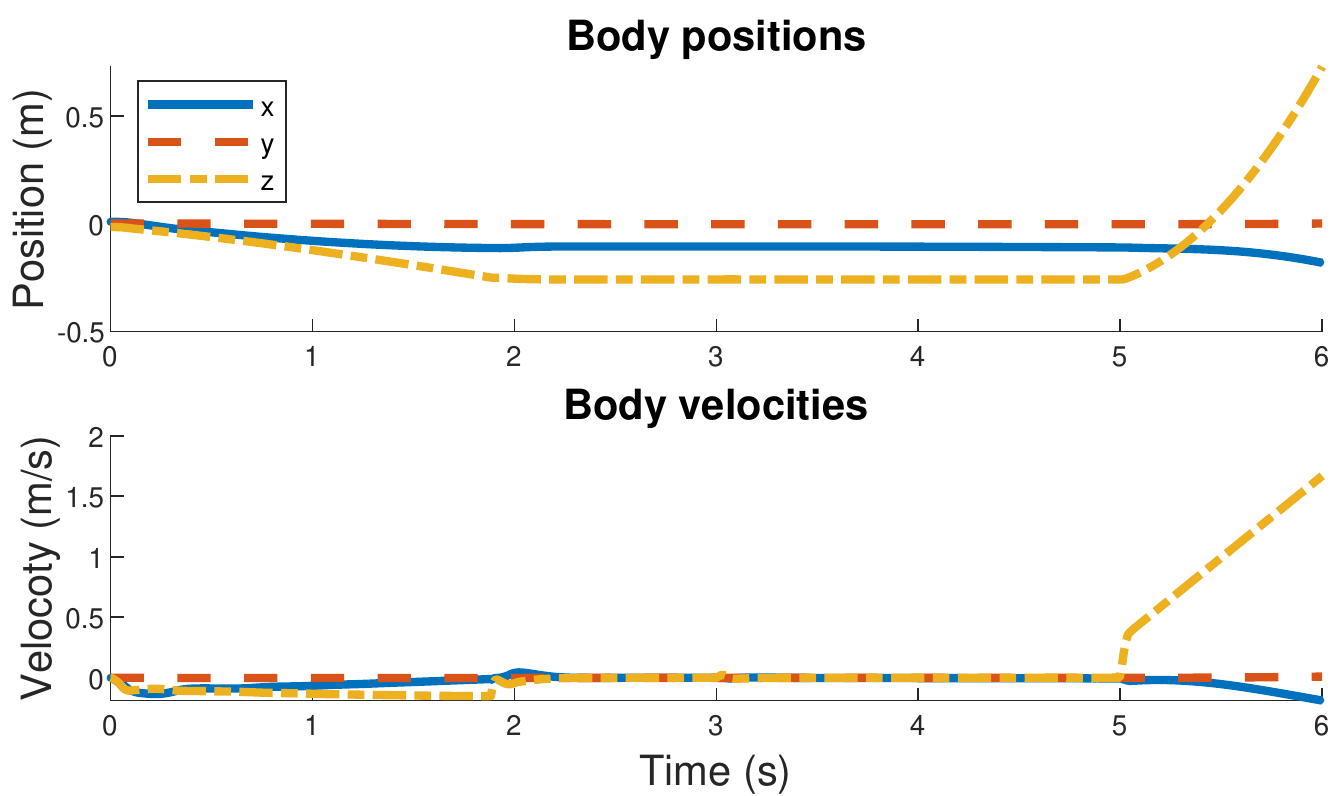}
    \caption{Position and velocity of the body during morphing. }
    \label{fig: State during morphing}
\end{figure}

A simulation study was performed to show its transforming capability from the legged to the UAV configuration. The joint trajectories of all the joints are plotted as shown in Fig. \ref{fig: trajectories}. The joint reference tracking performance is great, though there is a slight tracking delay which is caused by the computational time of the discrete to physical signal transformation. In Fig. \ref{fig: State during morphing} we can see the body position during the morphing process of the robot. The robot started to crouch and the perch touched the ground at the 2 second mark, which was then followed by the leg expanding to transform into the UAV configuration. Then the robot began to fly using the propellers and lifted off the ground.
\newpage

%% Modeling
% \chapter{Dynamic Modeling of Two-Contact-Point Trotting}
\chapter{Husky Carbon Modeling}
\label{chap:modeling}

\section{Husky Reduced-Order Model (HROM)}

\begin{figure}[h!]
\centering
\includegraphics[scale=0.7]{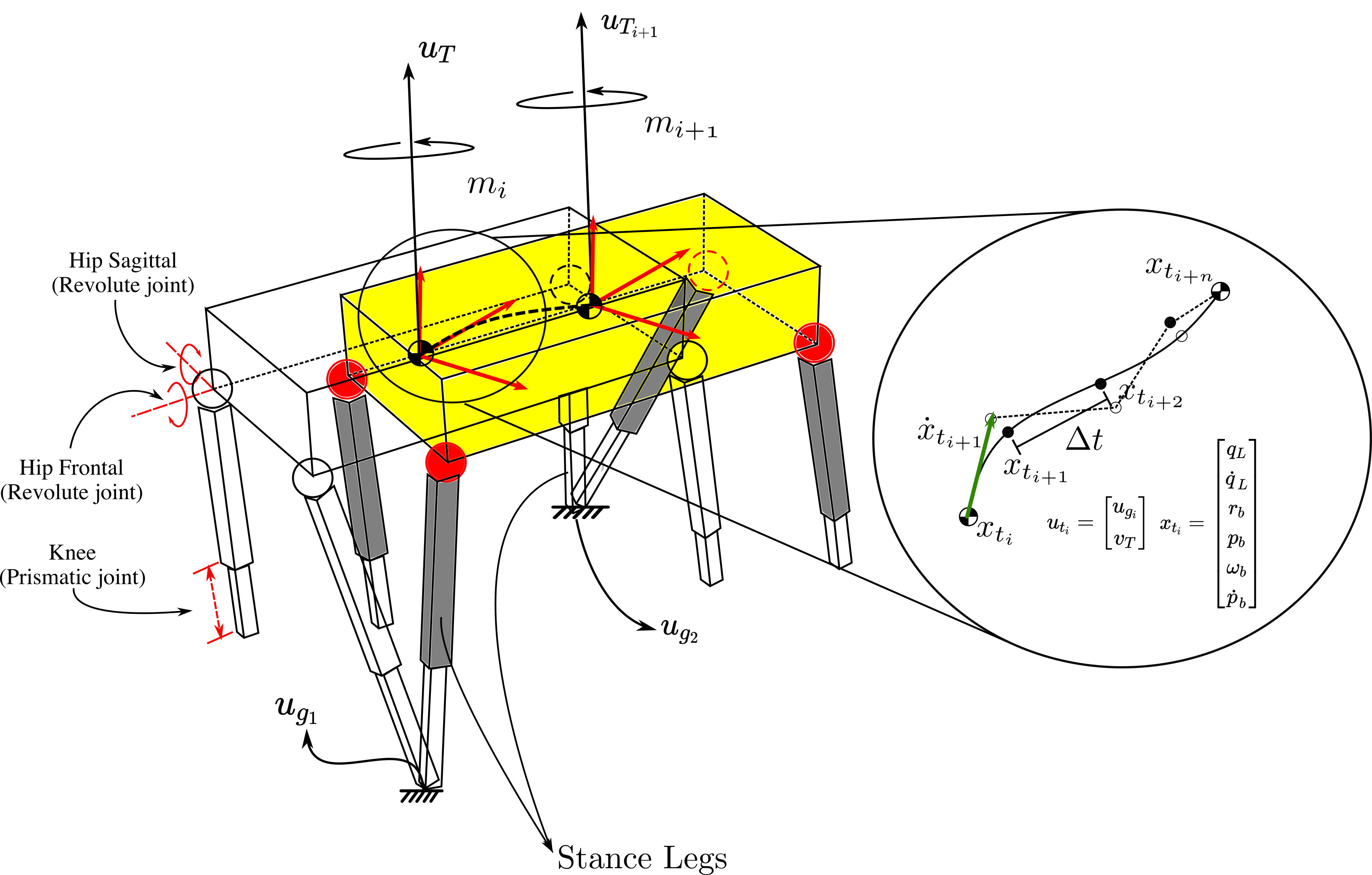}
\caption{Illustrates Husky Reduced-Order Model (HROM)}
\label{fig:hrom}
\end{figure}

\begin{figure*}
    \centering
    \includegraphics[width=1\linewidth]{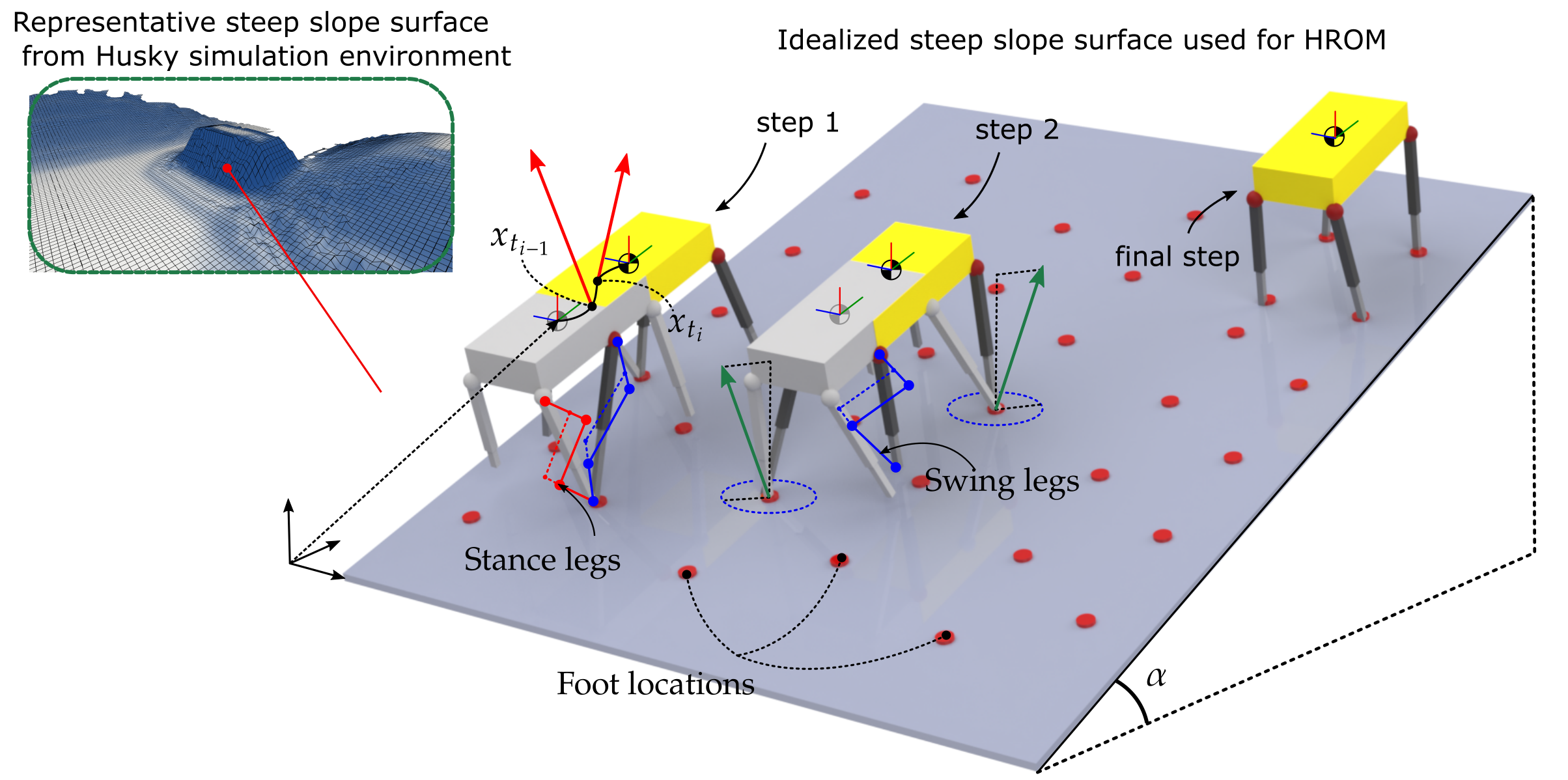}
    \caption{Shows the HROM used to formulate the control design for the WAIR problem.}
    \label{fig:fbd}
\end{figure*}

We use a reduced-order model, called HROM, for the collocation-based control design. In the HROM, each leg is assumed to be massless, that is, all masses are incorporated into the body yielding a 6-DOF model representing the torso's linear and orientation dynamics. Each leg is modeled using two hip angles (frontal and sagittal) and a prismatic joint to describe the leg end position. In HROM, the thruster forces are applied to the body center of mass (COM) and the ground reaction forces (GRF) are applied at the foot end positions. 

Let the superscript $b$ represent a vector defined in the body frame (e.g., $a^b$), and the rotation matrix $R_b\in SO(3)$ represents the rotation of a vector from the body frame to the inertial frame (e.g., $a = R_b a^b$). As such, the foot-end positions in the HROM can be derived using the following kinematics equations:
\begin{equation}
\begin{aligned}
    p_{f_i} &= p_b + R_b l_{h_i}^b + R_b l_{f_i}^b \\
    l_{f_i}^b &= R_{y}\left(\phi_{i}\right) R_{x}\left(\gamma_{i}\right)
    \begin{bmatrix} 
    0, & 0, & -r_{i}
\end{bmatrix}^\top,
\label{eq:foot_pos}
\end{aligned}
\end{equation}
where $p_{f_i}$ and $p_b$ denote the world-frame position of leg-ends and body position, respectively. $l_{h_i}^b$ and $l_{f_i}^b$ denote the body-frame position of hip-COM and foot-hip, respectively. $R_{x}$ and $R_{y}$ denote the rotation matrices around the x- and y-axes. Finally, $\phi_i$ and $\gamma_i$ are the hip frontal and sagittal angles respectively, and $r_i$ is the prismatic joint length.

Let $\omega_b$ be the body angular velocity vector in the body frame and $g$ denote the gravitational acceleration vector. The legs of HROM are massless, so we can ignore all leg states and directly calculate the total kinetic energy $\mathcal{K}= \frac{1}{2}m \dot p_b^{\top} \dot p_b + \frac{1}{2}\omega_b^{\top} J_b \omega_b$ (where $m$ and $J_b$ denote total body mass and mass moment of inertia tensor). The total potential energy of HROM is given by $\mathcal{V}=-m p_b^{\top} g$. Then, the Lagrangian $\mathcal{L}$ of the system can be calculated as $\mathcal{L} = \mathcal{K} - \mathcal{V}$. Hence, the dynamical equations of motion are derived using the Euler-Lagrangian formalism. 

The body orientation is defined using Hamilton's principle of virtual work and the modified Lagrangian for rotation dynamics in SO(3) to avoid using Euler rotations which can become singular during the simulation. The equations of motion for HROM are given by
\begin{equation}
\begin{gathered}
    \textstyle \frac{d}{d t} \left( \frac{ \partial \mathcal{L}}{\partial \dot p_b } \right ) - \frac{\partial \mathcal{L}}{\partial p_b} = f_{gen}, \qquad
    \dot{R}_b = R_b\,[\omega_b]_\times \\
    \textstyle \frac{d}{dt}\left( \frac{\partial \mathcal{L}}{\partial \omega_b}  \right) + 
    \omega_b \times \frac{\partial \mathcal{L}}{\partial \omega_b} + 
    \sum_{j=1}^{3} r_{b_j} \times \frac{\partial \mathcal{L}}{\partial r_{b_j}} = \tau_{gen},
\end{gathered}
\label{eq:euler-lagrangian}
\end{equation}
where $f_{gen}$ and $\tau_{gen}$ are the the generalized forces and moments (from GRF and thrusters), $[\,\cdot\,]_\times$ is the skew operator, and $R_b^\top = [r_{b_1}, r_{b_2}, r_{b_3}]$ (i.e., $r_{b_j}$ are the columns of $R_b$). The HROM model can then be solved from into the following standard form:
\begin{equation}
M\,
\begin{bmatrix}
\ddot p_b \\ \dot \omega_b
\end{bmatrix} 
+ H = \textstyle \sum_{i=1}^{2} B_{g_i}\, u_{g_i} + B_T\, u_T,
\label{eq:eom_dynamics}
\end{equation}
where $M$ is the mass-inertia matrix, $H$ contains the Coriolis matrix and gravity vector, $B_{g_i} u_{g_i}$ represent the generalized force due to the GRF $u_{g_i}$ acting on the i-th foot where $B_{g_i} = \Big(\frac{\partial \dot p_{f_i}}{\partial v}\Big)^\top$ and $v = [\dot p_b^\top, \omega_b^\top]^\top$. In Eq.~\ref{eq:eom_dynamics}, $u_T$ denotes the thruster action and $B_T= \Big(\frac{\partial \dot p_b}{\partial v}\Big)^\top$.

Note that the matrix $M$, $H$, $B_{g_i}$, and $B_T$ are functions of the robot's posture, that is, the leg joint variables $\phi_i$, $\gamma_i$, and $r_i$. The way we control posture is achieved by finding the proper accelerations for the desired body motion which is addressed by the collocation-based method as discussed later. The joint states and the inputs are defined as follows
\begin{equation}
    q_L = [\dots,\phi_i, \gamma_i, r_i,\dots]^\top, \qquad
    \ddot q_L = u_L
\end{equation}
Combining both the body dynamics given by Eq.~\ref{eq:eom_dynamics} and massless-leg states given above form the HROM state vector which is given by:
\begin{equation}
    x = [q_L^\top, \dot q_L^\top, r_b^\top, p_b^\top, \omega_b^\top, \dot p_b^\top]^\top,
\end{equation}
where the $r_b$ denotes the columns of the rotation matrix $R_b$. The state-space model of HROM can be defined as 
\begin{equation}
    \dot x = f_{_{ROM}}(x, u_L, u_g, u_T)
    \label{eq:hrom-state-space}
\end{equation}
which will be utilized for WAIR control of the high fidelity model in SimScape.

\section{HROM Ground Contact Model}

The ground model used in the HROM is fined tuned to match the model in the SimScape model. This section, briefly explains the HROM ground contact dynamics. The ground model used in HROM is given by 
\begin{equation}
\begin{aligned}
    u_{g_i} &= \begin{cases} \, 0 ~~  \mbox{if } p_{f_{i,z}} > 0  \\
     [u_{g_{i,x}},\, u_{g_{i,y}},\, u_{g_{i,z}}]^\top ~~ \mbox{else} \end{cases} \\
    u_{g_{i,z}} &= -k_1 p_{f_{i,z}} - k_{2} \dot p_{f_{i,z}} \\
    u_{g_{i,j}} &= - s_{j} u_{g_{i,z}} \, \mathrm{sgn}(\dot p_{f_{i,j}}) - \mu_v \dot p_{f_{i,j}} ~~  \mbox{if} ~~j=x, y\\
    s_{j} &= \Big(\mu_c - (\mu_c - \mu_s) \mathrm{exp} \left(\frac{-|\dot p_{f_{i,j}}|^2}{v_s^2}  \right) \Big)
\end{aligned}
\end{equation}
\noindent where $p_{f_{i,j}},~~j=x,y,z$ are the $x-y-z$ positions of the contact point; $u_{g_{i,j}},~~i=x,y,z$ are the $x-y-z$ components of the ground reaction force assuming a point contact takes place between the robot and the ground substrate; $k_{1}$ and $k_{2}$ are the spring and damping coefficients of the compliant surface model; $\mu_c$, $\mu_s$, and $\mu_v$ are the Coulomb, static, and viscous friction coefficients; and, $v_s > 0$ is the Stribeck velocity.

\chapter{Control Design}

Flight control and trotting control in Husky were reported in previous works. This chapter presents a new control design for the WAIR maneuver. The specific contribution of this Thesis regarding WAIR simulations lies in creating the high-fidelity model that is used for WAIR control algorithm evaluations. Therefore, while not contributed by this MS work, this control chapter is presented to provide a complete picture of the Thesis contribution. 

\section{Collocation based Control}
Consider $N$ time intervals during the WAIR maneuver 
\begin{equation}
0=t_1<t_2<\ldots<t_N=t_f
    \label{eq:?}
\end{equation}
\noindent where $t_k$ denotes discrete times. First, we discretize the continuous model given by Eq.~\ref{eq:hrom-state-space} using an explicit Euler integration scheme as follows
\begin{equation}
\begin{aligned}
    x_{k+1} &= x_{k} + \Delta t f_{_{ROM}} (x_{k},u_{L,k}, u_{g,k}, u_{T,k}) , \\
    &\quad k=1, \ldots, N, \quad 0 \leq t_k \leq t_f
\end{aligned} 
\label{eq:discrete_model}
\end{equation}   
where $\Delta t$ is the integration time step, $t_k$ is the discrete time at k-th discrete step, $x_k = x(t_k)$ denotes the state vector at $t_k$, and $u_{i,k} = u_{i,k}(t_k)$ (where $i=L,~g,~T$) is the input. Let $x_{r,k}$ be the state reference and $e_k = x_{r,k} - x_{k}$ be the tracking error for the states of HROM. Then, we consider the following cost function 
\begin{equation}
\begin{aligned}
    J = \sum_{k=1}^{N} \left( e_k^\top \,Q\, e_k + u_{k-1}^\top \,R\, u_{k-1} \right)
    % \min_{\vect{u}_k} \quad &
    % J = \sum_{k=1}^{n_h} \left( \vect{e}_k^\top \,\vect{Q}\, \vect{e}_k + \vect{u}_{k-1}^\top \,\vect{R}\, \vect{u}_{k-1} \right) \\
    % \textrm{s.t.} \quad &
    % \vect{x}_{k+1} = \vect{x}_{k} + \Delta t \, \left( \vect f (\vect{x}_k) + \vect g (\vect{x}_k)\, \vect{u}_k \right) \\
    % & \vect{h}_c(\vect{x}_{k}, \vect{u}_k) < \vect{0}, \quad \textrm{ for } k \in \{0,\dots, n_h-1\}\\
    % &  \vect{u}_{min} \leq \vect{u}_k \leq \vect{u}_{max}
\end{aligned} 
\label{eq:cost}
\end{equation}   
where $N$ is ..., $Q$ and $R$ denote the diagonal weighting matrices used to penalize the tracking performance and control efforts, respectively. Note that from here onward as shown in Eq.~\ref{eq:cost}, all inputs, including the GRF, joint accelerations, and thruster forces, are stacked in the input vector $u$.

% cubic collocation at Lobatto points
Our objective is to find $u_k$ based on cubic collocation at Lobatto points such that $J$ is minimized during the WAIR maneuver. We consider 2N boundary conditions given by
\begin{equation}
    \begin{aligned}
        % r_i\left(x(0), x\left(t_f\right), t_f\right)=&0, \quad i=1, \ldots, k \leq 2 N
        r_i\left(x(0), x\left(t_f\right), t_f\right)=&0
        % &?=?\\
        % &\vdots\\
        % &?=?
    \end{aligned}
\end{equation}
\noindent to enforce continuity of the state vector evolution. We consider N inequality constraints given by
\begin{equation}
    % g_i(x(t), u(t), t) \geq 0, \quad i=1, \ldots, m, \quad 0 \leq t \leq t_f
    g_i(x(t), u(t), t) \geq 0
\end{equation}
\noindent to limit the input $u_k$ in each discrete time period. We stack all of the discrete states $x_k$ and inputs $u_k$ from the HROM model given by Eq.~\ref{eq:discrete_model} in the vectors $X = \left[x^\top_1, \ldots, x^\top_k\right]^\top$ and $U = \left[u^\top_1, \ldots, u^\top_k\right]^\top$. In this work, we resolve the optimal solutions for the WAIR maneuver for fixed $t_f$. However, it is possible $t_f$, as the decision parameter in the optimization problem, that is, we add final discrete time $t_f$ as the last entry of the decision parameter vector $Y$,
\begin{equation}
Y=\left(X,U, t_f\right) \in \mathbb{R}^{2N+1}
    \label{?}
\end{equation}
We find $X$ and $U$ as the decision parameters of the optimization problem using MATLAB's nonlinear optimization toolbox. To resolve the optimization problem rapidly, we employ an interpolation approach to approximate $x_k$ and $u_k$. We take input to be as the linear interpolation function between $u(t_k)$ and $u(t_{k+1})$ for $t_k \leq t<t_{k+1}$, that is, $u_{int}$ is given by
\begin{equation}
u_{int}(t)=u\left(t_k\right)+\frac{t-t_k}{\Delta t}\Big(u\left(t_{k+1}\right)-u\left(t_k\right)\Big)
    \label{}
\end{equation}
\noindent In addition, we interpolate the HROM's states $x(t_k)$ and $x(t_{k+1})$ too. This way, the speed of control computations increases considerably. We take a nonlinear cubic interpolation which is continuously differentiable with 
\begin{equation}
    \dot{x}_{\mathrm{int}}(s)=f_{_{ROM}}(x(s), u(s), s)
\end{equation}
at $s=t_k$ and $s=t_{k+1}$. To do this, we write the following system of equations:
\begin{equation}
    \begin{aligned}
x_{int}(t) &=\sum_{k=0}^3 c_k^j\left(\frac{t-t_j}{h_j}\right)^k, \quad t_j \leq t<t_{j+1}, \\
c_0^j &=x\left(t_j\right), \\
c_1^j &=h_j f_j, \\
c_2^j &=-3 x\left(t_j\right)-2 h_j f_j+3 x\left(t_{j+1}\right)-h_j f_{j+1}, \\
c_3^j &=2 x\left(t_j\right)+h_j f_j-2 x\left(t_{j+1}\right)+h_j f_{j+1}, \\
\text { \textbf{where} } f_j &=f_{_{ROM}}\left(x\left(t_j\right), u\left(t_j\right), t_j\right),\\
h_j&=t_{j+1}-t_j .
\end{aligned}
\label{eq:cubic-lobatto}
\end{equation}
The interpolation function used for $x_{int}$ must satisfy the derivatives at the discrete points and at the middle of sample times $t_{c,i}$. 

By inspecting Eq.~\ref{eq:cubic-lobatto}, it can be seen that the derivative terms at the boundaries $t_{i}$ and $t_{i+1}$ are satisfied. Therefore, the only remaining constraints in the nonlinear programming constitute the collocation constraints at the middle of $t_i-t_{i+1}$ time interval, the inequality constraints at $t_i$, and the constraints at $t_1$ and $t_f$. This property of the interpolation method given by Eq.~\ref{eq:cubic-lobatto} reduces the total number of equations that must be resolved yielding a considerable speed in control computations. Hence, the remaining constraints in the nonlinear programming are given by:
\begin{equation}
\begin{aligned}
f_{_{ROM}}\Big(x_{int}\left(t_{c,i}\right), u_{int}\left(t_{c,i}\right)\Big)-\dot x_{int}\left(t_{c,i}\right)&=0\\
g\Big(x_{int}\left(t_i\right), u_{int}\left(t_i\right), t_i\Big) &\geq 0\\
r\Big(x_{int}\left(t_1\right), x_{int}\left(t_N\right), t_N\Big)&=0
\end{aligned}
    \label{eq:fdc-controller}
\end{equation}
\noindent Given that the computational structure is spatially discrete with large costs associated with its curse of dimensionality, this collocation scheme results in a smaller number of parameters for interpolation polynomials which enhance the computation performance. We resolve this optimization problem using MATLAB fmincon function.

%% Simulation Results
% intro.tex:

\chapter{SimScape Simulation Results}
\label{chap:results}

In order to test the behavior of Husky, simulation and physical tests were conducted using MATLAB, Simscape and Simulink Realtime. MATLAB allowed the modeling and simulation of the expected controller and behavior of the system. Simscape allowed the urdf generated from the CAD model of the system to be incorporated for simulation in a physics environment to test the expected behavior of the physical system.

\section{SimScape Husky Model}
\label{sec:simscape-husky}
\begin{figure}
    \centering
    \includegraphics[width = 1.0 \linewidth]{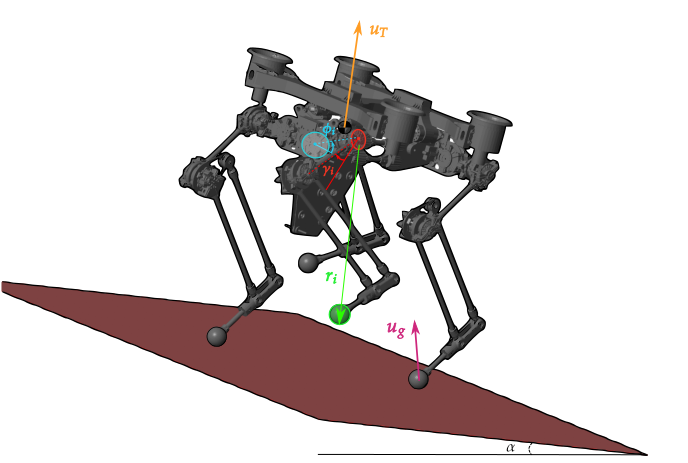}
    \caption{Shows the URDF model and SimScape environment developed to realistically evaluate Husky performance.}
    \label{fig:simscape}
\end{figure}

We created a high fidelity model of Husky WAIR model in MATLAB SimScape (see Fig.\ref{fig:simscape}). This model has a total number of 18 degress of freedom (DOF). In this model, we consider a total number of 13 distributed mass elements located at the main body, hips, thighs, and shins. The main body possesses a total mass of 5.0 kg and the diagonal components of mass moment of inertia, $I_{xx}$, $I_{yy}$, $I_{zz}$ ($kg.m^{2}$) $[0.0981867, 0.0844185, 0.164599]$. The hips, thighs, and shins possess body mass of 0.220, 0.060, 0.050 kg. The forward and inverse kinematics of the robot is erected based on the body coordinate frames shown in Fig.~\ref{fig:simscape}. The main body coordinate frame and world frame are linked together using Euler angles.

The sloped surface considered in our analyses is flat with a known slope angle. The ground contact model used in the high-fidelity model is a smooth spring-damper model for normal force, with stiffness $100 N/m$, damping $1 \times 10^{3} N/(m/s)$ and transition region width of $1 \times 10^{-3} m$. And for tangential force, a smooth stick-slip model is used with a coefficient of static friction of $1.8$ and a coefficient of Dynamic Friction of $1.0$.

\section{Ground Model}
\label{sec:simscape-ground}

In order to incorporate a ground model in simulation, a world plane was created and the SimScape multibody contact forces library was utilized. The library contains contact force models and force/friction laws to evaluate external mechanics and effects. The contact was defined as a sphere-plane contact with linear force law and stick-slip continuous friction law. The library also allows us to define the contact stiffness/damping and coefficients of friction. Furthermore, it allows checking for impact of the leg ends to coordinate gait motions by monitoring the normal force values throughout the simulation.

\section{Simulation Results and Discussion}
\label{sec:}

\begin{figure}[h]
    \centering
    \includegraphics[scale=0.4]{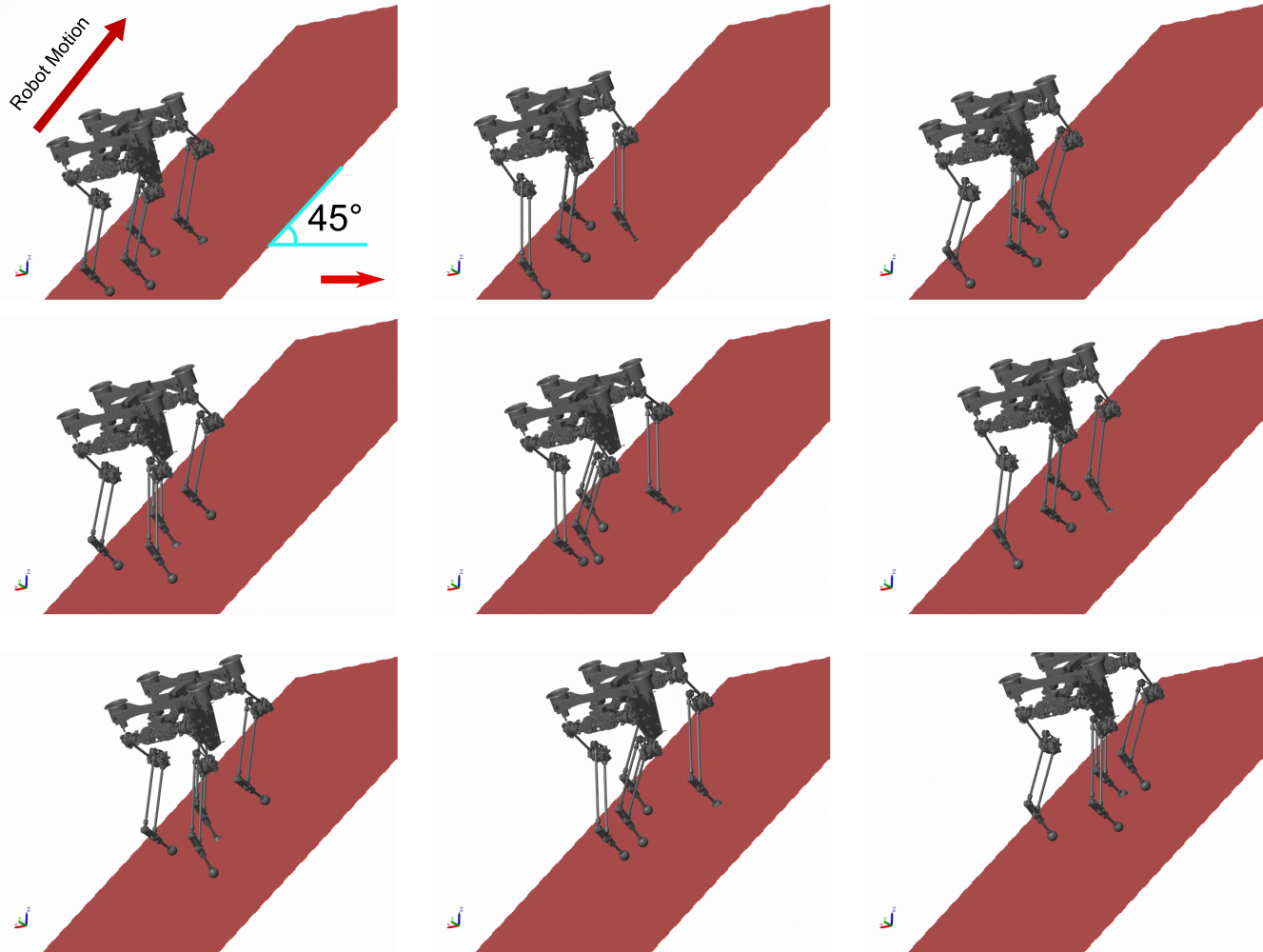}
    \caption{Simulated snapshots of the WAIR maneuver on a 45-deg slope using Husky's high-fidelity model in MATLAB SimScape.}
    \label{fig:wair_sim}
\end{figure}

\begin{figure}
    \centering
    \includegraphics[width=\linewidth]{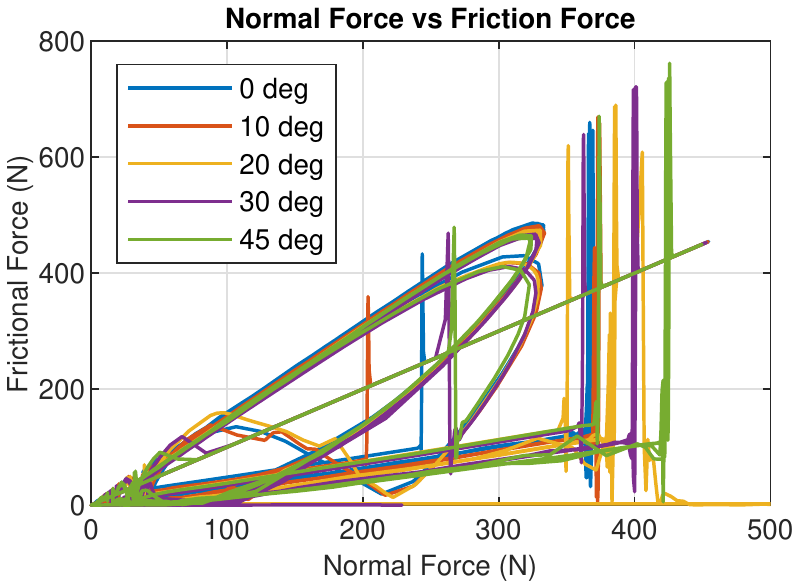}
    \caption{Illustrates the normal force $u_{g_{i,z}}$ versus tangential $\sqrt{u^2_{g_{i,x}}+u^2_{g_{i,y}}}$ at the stance leg-ends during the WAIR maneuver.}
    \label{fig:grf2}
\end{figure}

\begin{figure}
    \centering
    \includegraphics[width=\linewidth]{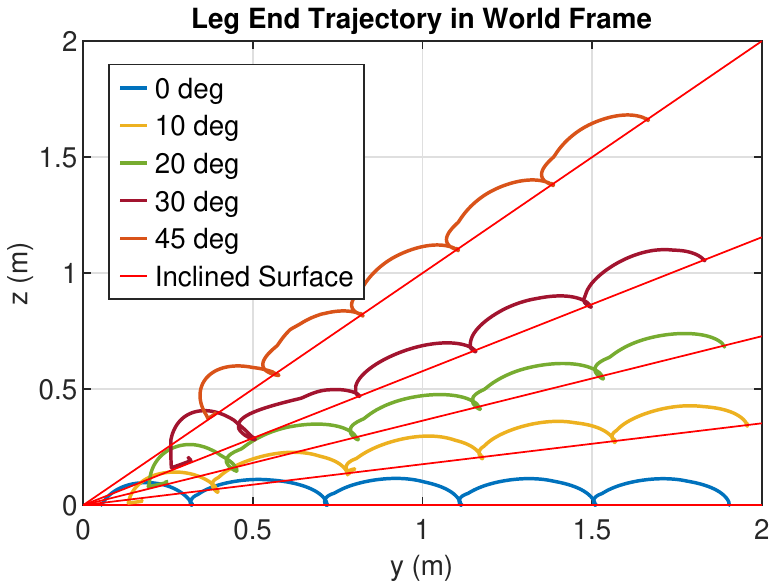}
    \caption{Show the swing leg-end trajectory during slope climbing. The dashed red line denotes the slope surface.}
    \label{fig:leg_end_pos}
\end{figure}

\begin{figure}
    \centering
    \includegraphics[width=\linewidth]{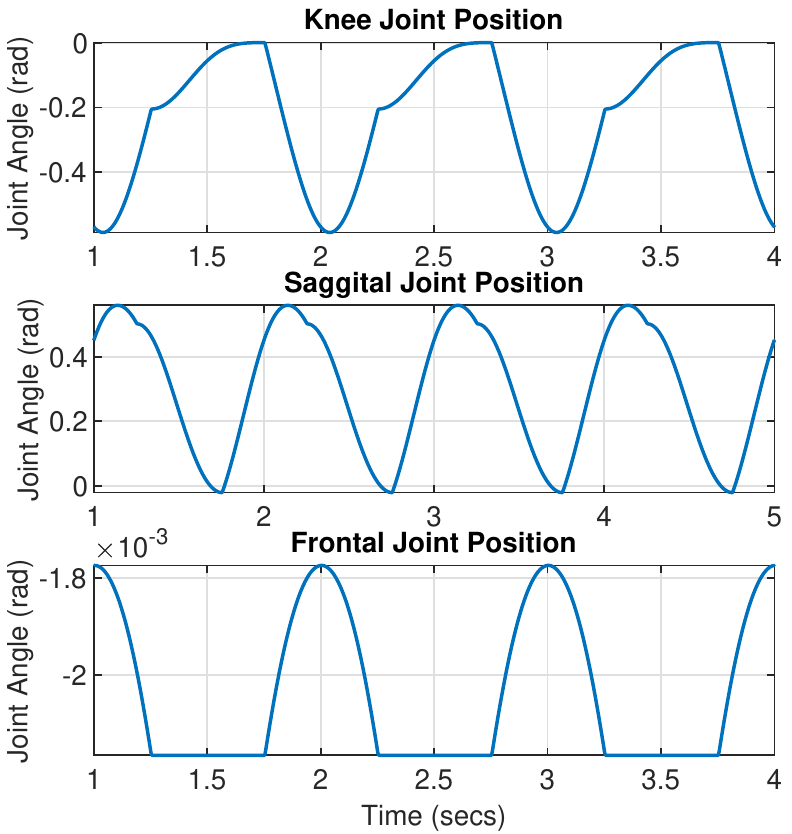}
    \caption{Show the joint angles for the front leg during slope climbing.}
    \label{fig:joint_position}
\end{figure}

\begin{figure}
    \centering
    \includegraphics[width=\linewidth]{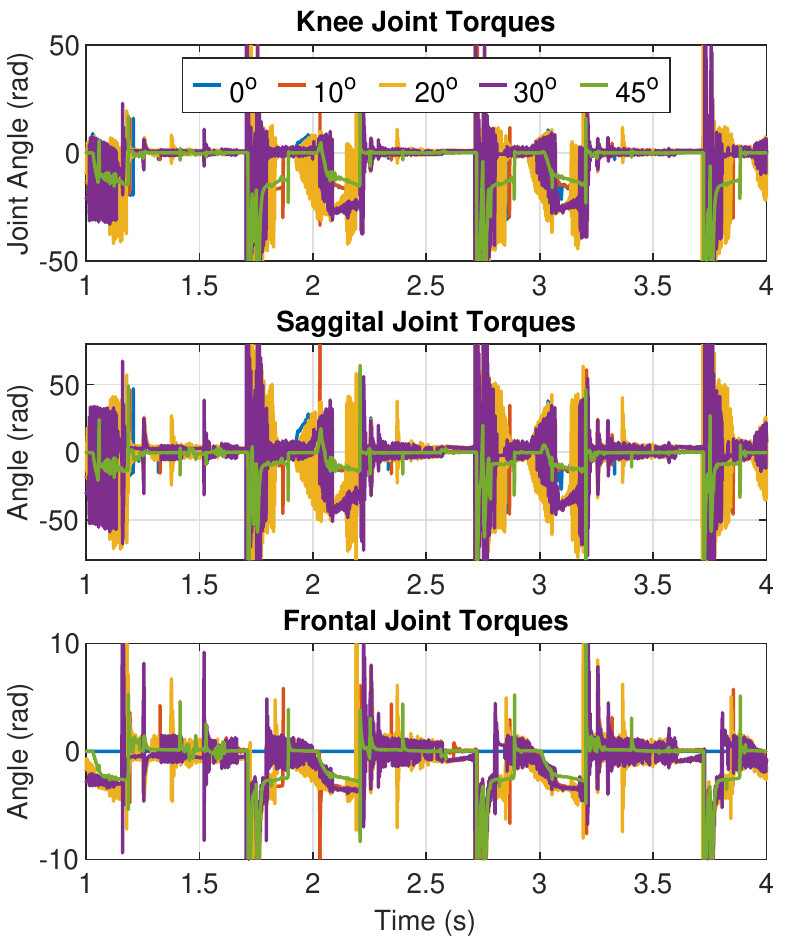}
    \caption{Illustrates predicted joint torques}
    \label{fig:sim-torque}
\end{figure}

Here, we briefly report these simulation results and discuss the lessons learned.\\
We simulated the closed-loop WAIR problem. Figure~\ref{fig:wair_sim} shows the snapshots of the high-fidelity SimScape model. In this simulation, Husky walks up a 45-deg incline. Other WAIR simulations include walking over 0 deg, 10 deg, 20 deg, and 30 deg inclinations. The gait remained fixed in these simulations, as shown in Figure \ref{fig:joint_position}. Figure~\ref{fig:leg_end_pos} shows the swing leg-end positions for different inclinations. The manipulated ground contact forces for different inclination scenario satisfies friction cone conditions shown in Fig.~\ref{fig:grf2}. The joint torques, shown in Fig.~\ref{fig:sim-torque}, show higher peaks at lower inclinations as the legs carry a larger percentage of the total body weight and thruster actions contribute less. However, in higher inclinations, thruster contributions become larger.\\
Our simulation results covers kinematics modeling including foot placement and joint motion as well as dynamical analysis such as inertial force and GRF studies. The Husky URDF model was loaded into SimScape while allowing for the mass and inertial properties of the system to be incorporated in the model explorer. As a result, the SimScape model realistically predicts vehicles dynamics.\\
Through the use of transform sensors, body and joint position-orientation can be obtained. In addition, the SimScape multibody contact forces library offered analyses of the normal and frictional forces of the leg-end contacts with the modeled ground. The ground model parameters were set to reflect the surface the physical Husky robot was tested experiments. Since it is important to setup the simulation to resemble the experiments the ground model properties are set accordingly. As such, the contact stiffness and damping are set to $\num{5e-3}N/mm/mm^{2}$ and $\num{2e-10}N/mm/mm^{2}$, respectively. Kinetic and static friction are set to 0.7 each.\\

In Fig.~\ref{fig:sim-torque}, the actuation torques applied to the joints are shown. Simulation of the actuation torques necessary to produce the given gait allows us to obtain insight to the feasibility of the current actuator's capabilities to  achieve the desired motion. \\
Figure~\ref{fig:grf2} illustrates the ground reaction forces experienced throughout the gait. The ground reaction forces can help us determine whether friction cone constraints mentioned in Section 2.5.2 are satisfied. In order to avoid slippage, the ratio of the friction force to the normal force at any given moment must be be less than the coefficient of static friction, which was set to 0.7 in the simulation. We can see the moments each leg is in swing phase when the forces are at 0 and the when impact of leg end occurs (the prominent peaks).

%% Experimental Results
\chapter{Experiments}
\label{chapt:exp}
% Please talk about
% \begin{itemize}
%     \item PU tests 
%     \item Husky beta trotting tests
%     \item Husky tests with PU
% \end{itemize}
% This chapter aims to present the experiment results obtained from two distinct experiments,  Propulsion Unit (PU) test, and the thruster-assisted trotting test. 

In this chapter, I carried out two experiments: the PU test and the thruster-assisted trotting test. The PU test aimed to evaluate the performance of our Propulsion Unit without any load, providing an overall understanding of its capability. The purpose of the second test was to showcase the husky platform's ability to trot with the help of thrusters, so called "thruster-assisted locomotion", enabling both legged locomotion and aerial motion simultaneously.

\section{Propulsion Unit(PU) test}
\begin{figure}[h!]
    \centering
    \includegraphics[scale=0.5]{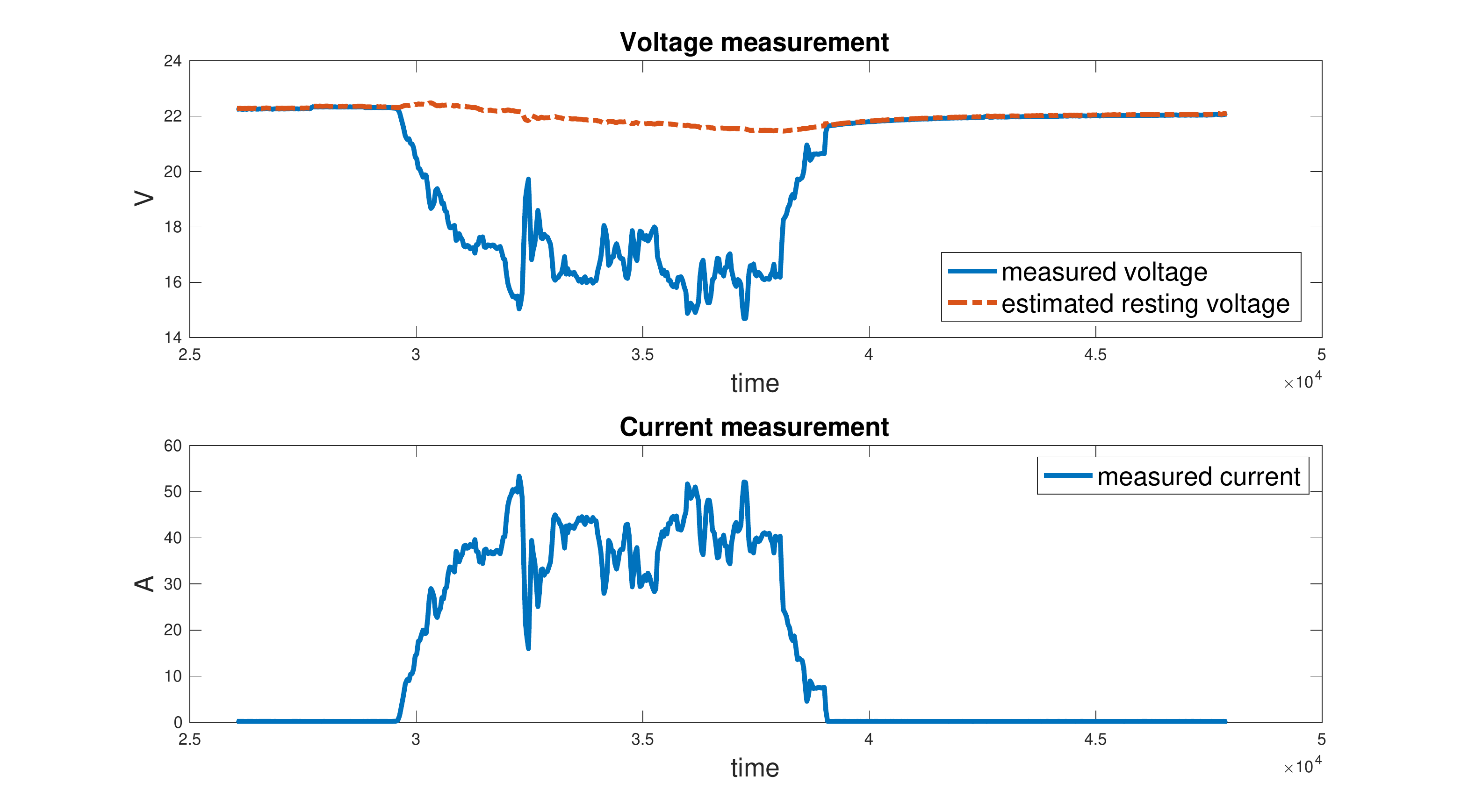}
    \caption{Show the voltage and current information during test of Propulsion Unit.}
    \label{fig:pu_voltage_current}
\end{figure}

In order to verify the functionality of the propulsion unit, a series of tests were conducted employing a movable gantry, which was connected to the unit via two ropes to mitigate the possibility of operational errors leading to damage. During the testing process, a joystick was used by the operator to control the system, while control commands in the form of Sbus signals were transmitted from a receiver located on the system side. The onboard ARM Cortex-M3 controller of the Pixhawk flight controller processed this information and transmitted it to the electric speed controller (ESC) in the form of PWM signals, allowing for adjustment of the current output to control each EDF. The results of the testing process are presented as follows:

The occurrence of a substantial voltage drop and an increase in the disparity between the desired and measured voltages~\ref{fig:pu_voltage_current} can be taken as an indication that the system has been energized, a fact which is corroborated by the observed pattern in the current diagram.

\newpage   

% \begin{figure}[h!]
%     \centering
%     \includegraphics[scale=0.5]{fig/PU_data_no_trotting/Current_no_trotting.png}
%     \caption{Show the Current information during test of Propulsion Unit.}
%     \label{fig:pu_current}
% \end{figure}

\begin{figure}[h!]
    \centering
    \includegraphics[scale=0.43]{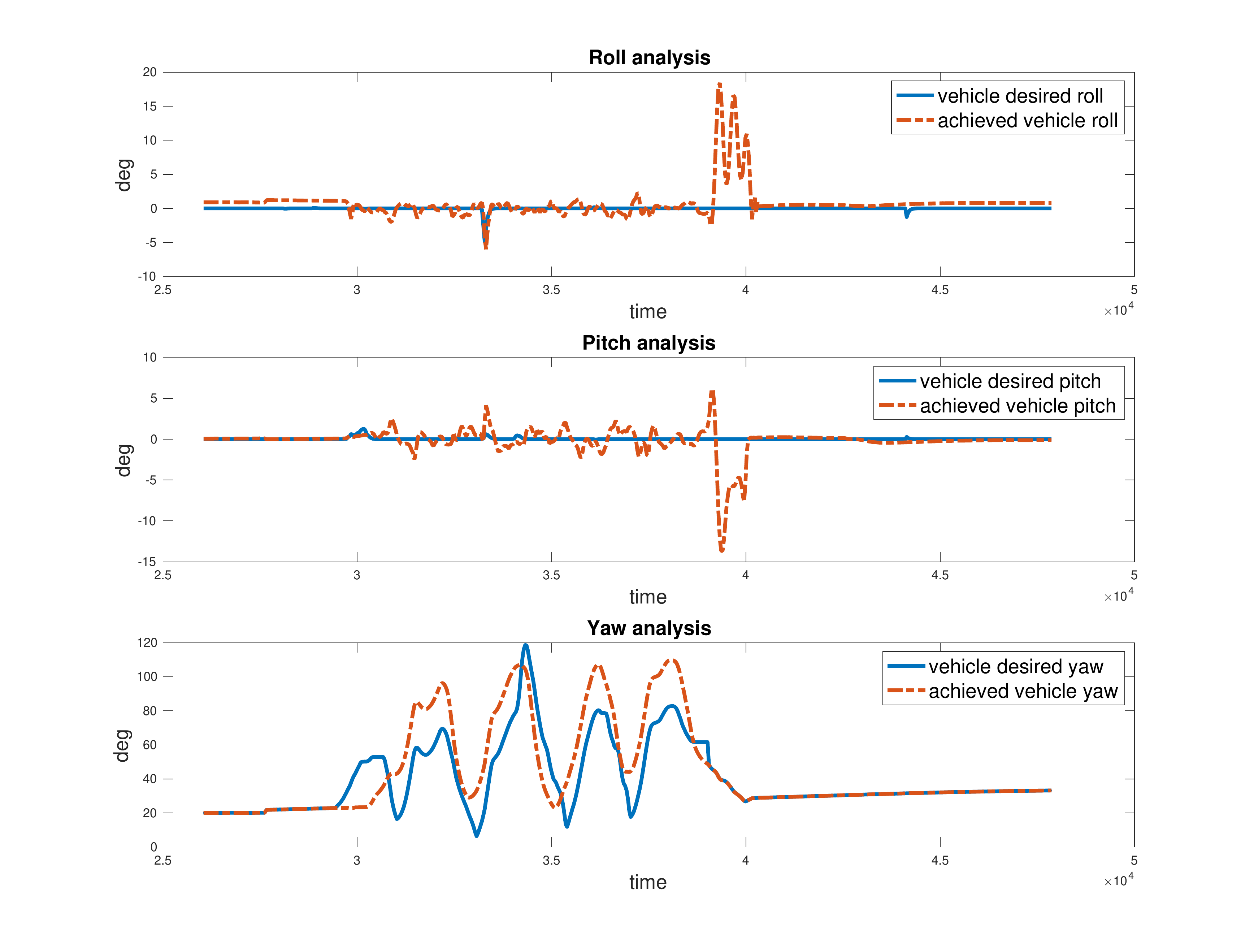}
    \caption{Show the orientation information during test of Propulsion Unit.}
    \label{fig:pu_no_tortting}
\end{figure}

The three subsequent diagrams illustrate that the measured state closely resembles the desired state, indicating that the feedback loop is responsive. However, the pitch and roll diagram shows some errors that result from the system reaching the rope length limit~\ref{fig:pu_no_tortting}. Additionally, when the robot is connected to the propulsion unit, external payloads may cause orientation errors and produce dissimilar outcomes.

% \begin{figure}[h!]
%     \centering
%     \includegraphics[scale=0.45]{fig/PU_data_no_trotting/Roll_no_trotting.png}
%     \caption{Show the orientation(roll) information during test of Propulsion Unit.}
%     \label{fig:pu_roll}
% \end{figure}

\clearpage

\section{PU test with Husky trotting in place}
\begin{figure}[ht!]
    \centering
    \noindent\makebox[\textwidth]{
    \includegraphics[scale=0.67]{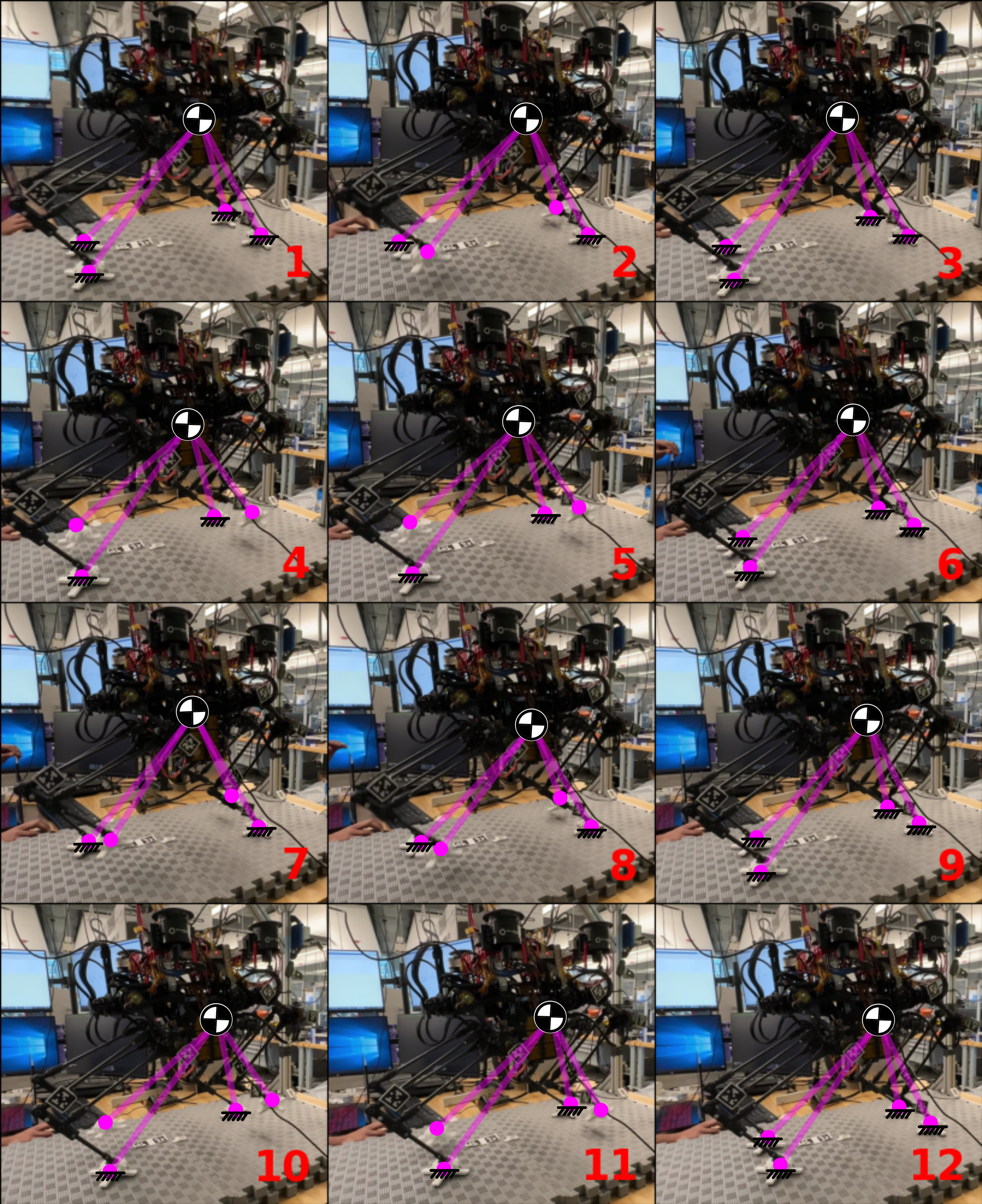}}
    \caption{The figure provided shows a discrete-time sequence from an experiment involving thruster-assisted trotting in place. The graph displays the observed patterns of legged locomotion during the trial, and the accompanying data specifies the thrust output from the thrusters used in the test.}
    \label{fig:husky_trotting}
\end{figure}
\newpage

\begin{figure}[h]
    \centering
    \includegraphics[scale=0.5]{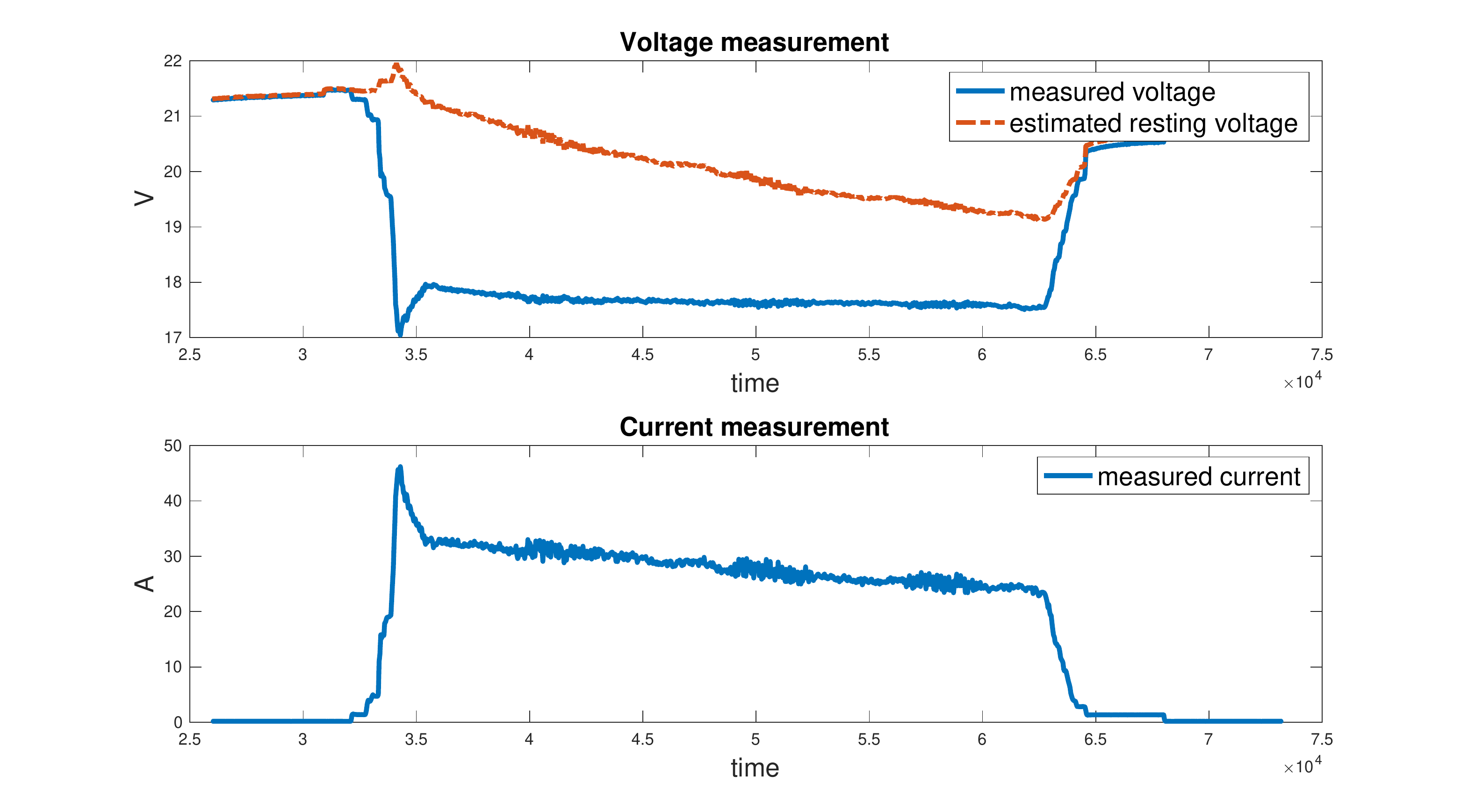}
    \caption{Show the voltage and current information during test.}
    \label{fig:pu_voltage_current_trotting}
\end{figure}

The initial two charts evince a remarkable similarity to the pattern witnessed during the Propulsion Unit (PU) test, depicting a discernible reduction in voltage~\ref{fig:pu_voltage_current_trotting} that corresponds to the engagement of the throttle, succeeded by a surge in current, as evidenced by the second chart. 

% \begin{figure}[h]
%     \centering
%     \includegraphics[scale=0.5]{fig/PU_data_trotting/Current_trotting.png}
%     \caption{Show the Current information during test.}
%     \label{fig:pu_current_trotting}
% \end{figure}
\newpage
Nonetheless, the following three graphs deviate significantly from this trend. Notably, the third to fifth graphs manifest a significant discrepancy between the intended and actual states, attributed to the increased payload(Husky Platform~\ref{fig:husky_trotting}). Additionally, these charts delineate the occurrence of three distinct high-frequency oscillations that connote the activation of the robot's trotting gait. It is worth noting that the amplitude of the roll is substantial, stemming from the robot's angular momentum.

\begin{figure*}[h!]
    \centering
    \includegraphics[scale=0.43]{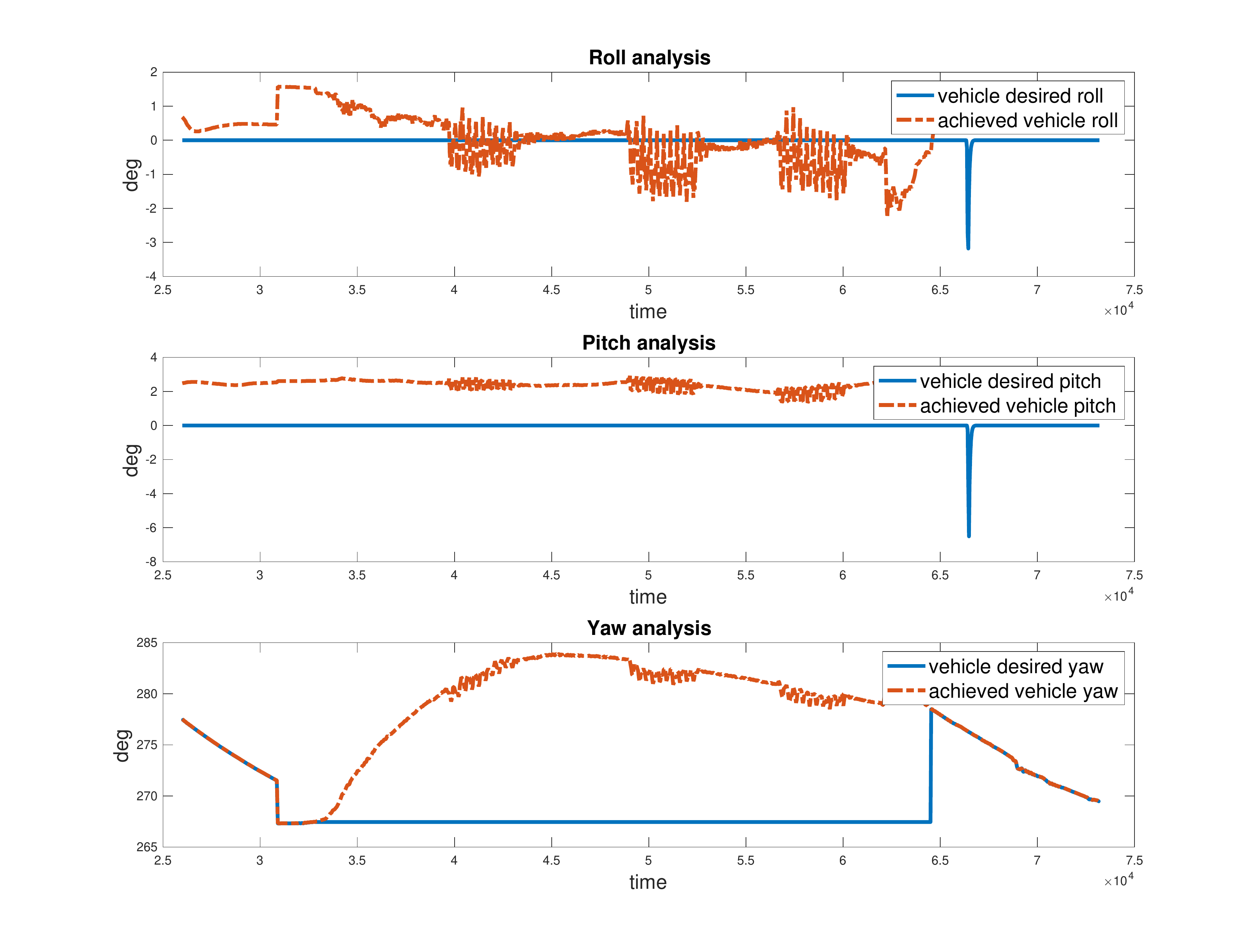}
    \caption{Show the orientation information during test.}
    \label{fig:pu_trotting}
\end{figure*}

% \begin{figure}[h!]
%     \centering
%     \includegraphics[scale=0.43]{fig/PU_data_trotting/Pitch_trotting.png}
%     \caption{Show the orientation(pitch) information during test.}
%     \label{fig:pu_pitch_trotting}
% \end{figure}
% \begin{figure}[h!]
%     \centering
%     \includegraphics[scale=0.43]{fig/PU_data_trotting/Roll_trotting.png}
%     \caption{Show the orientation(roll) information during test.}
%     \label{fig:pu_roll_trotting}
% \end{figure}

The following present charts exhibit the robot joint data collected during husky thruster-assisted trotting in place. The plotted information encompasses three separate tests, each characterized by three oscillations. The data suggests that the hip-sagittal and knee joints closely adhere to the reference data, indicating the efficacy of the feedback system implemented in the robot. However, minor discrepancies, registering at less than 0.01 degrees, are discernible in the hip frontal joints. These variations are believed to be attributable to the presence of noise during the robot's motion.

% \begin{figure}[h]
%     \centering
%     \includegraphics[width=0.9\textwidth, left]{fig/joint_angle_trotting.eps}
%     \caption{Desired and actual joint angle.}
%     \label{fig:joint-angle-BR}
% \end{figure}

\begin{figure*}[ht]
    \centering
    \noindent\makebox[\textwidth]{\includegraphics[width=1.1\linewidth]{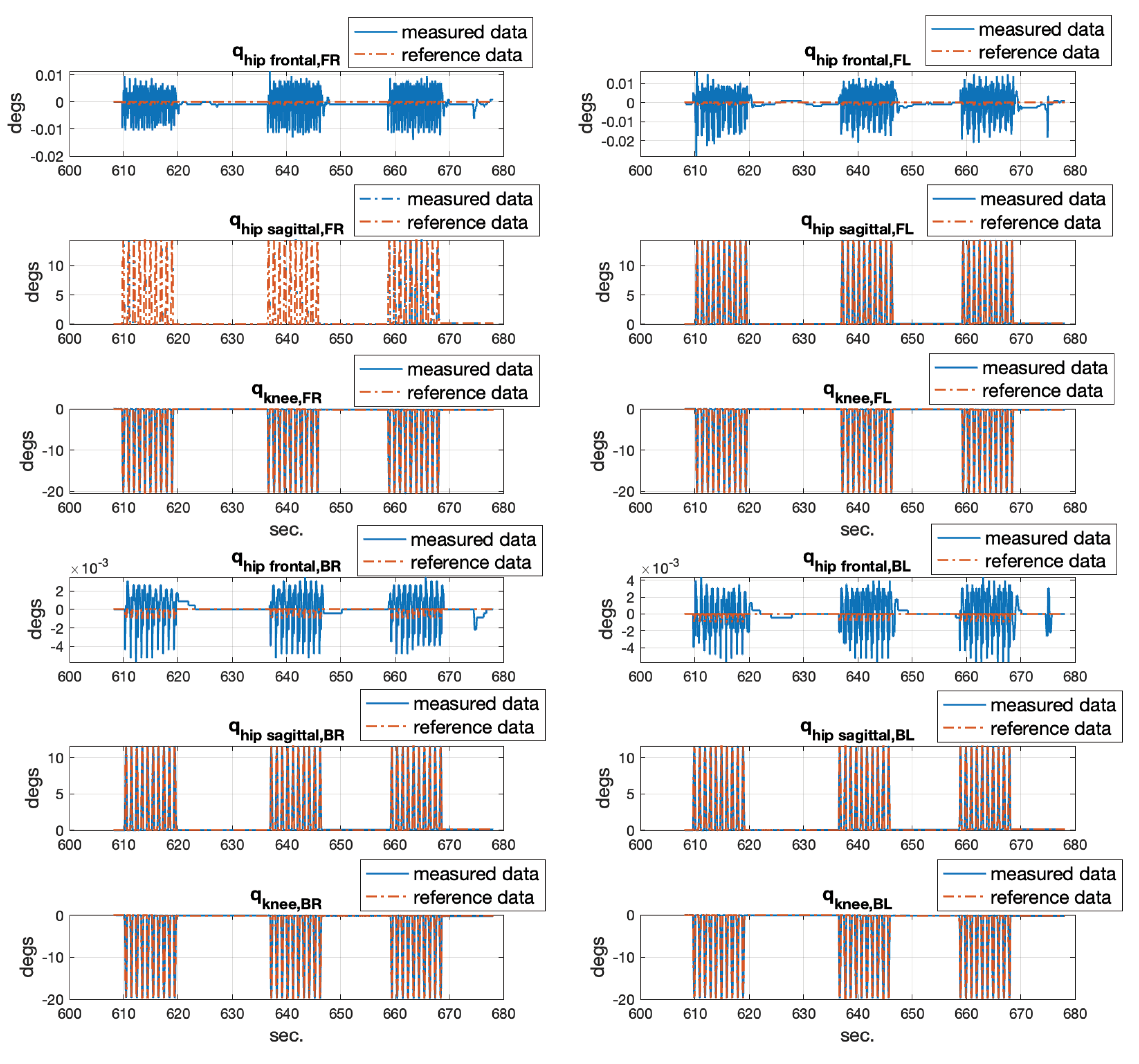}}
    \caption{Desired and actual joint angles during thruster-assisted trotting.}
    \label{fig:joint-angle}
\end{figure*}

% \begin{figure}[h]
%     % \centering
%     \includegraphics[scale=0.23]{fig/PU_data_trotting/JointAngleFL.png}
%     \caption{Desired and actual joint angle(FL).}
%     \label{fig:joint-angle-FL}
% \end{figure}

% \begin{figure}[h]
%     % \centering
%     \includegraphics[scale=0.23]{fig/PU_data_trotting/JointAngleFR.png}
%     \caption{Desired and actual joint angle(FR).}
%     \label{fig:joint-angle-FR}
% \end{figure}

%% conclusion
\chapter{Conclusion}
\label{chap:conclude}

The main contributions of this MS Thesis is centered around taking steps towards successful multi-modal demonstrations using Northeastern's legged-aerial robot, Husky Carbon. This work discusses the challenges involved in achieving multi-modal locomotion such as trotting-hovering and thruster-assisted incline walking and reports progress made towards overcoming these challenges. Animals like birds use a combination of legged and aerial mobility, as seen in Chukars’s wing-assisted incline running (WAIR), to achieve multi-modal locomotion. Chukars use forces generated by their flapping wings to manipulate ground contact forces and traverse steep slopes and overhangs. Husky's design takes inspirations from birds such as Chukars. This MS thesis outlines the mechanical and electrical details of Husky 's legged and aerial units. The thesis presents  simulated incline walking using a high-fidelity model of the Husky Carbon over steep slopes of up to 45 degrees.

% --- Bibliography ----
% \bibliographystyle{IEEEtran}  %'plain' for standard, 'unsrt' for correct order

% include bibliography definition
% \bibliography{bib/references}

% \bibliographystyle{unsrt}

\printbibliography

% --- Appendix ---
%\appendix
%\chapter{Simulink Model Sub-blocks}
%\input{tex/appendixa.tex}
%\chapter{Second Appendix Headline}
%%include anything you need in the appendix
%\include{appendix/appendix}

% --- Index ----
\printindex

% --- that's it ---
\end{document}